\documentclass[10pt,journal,compsoc]{IEEEtran}

\usepackage[UTF8,scheme=plain,heading=false,zihao=false,fontset=none]{ctex}

\usepackage{amsmath,amssymb,amsfonts}
\usepackage{amsthm}
\usepackage{mathtools}
\usepackage{bm}

\usepackage{xspace}
\usepackage[ruled,vlined,linesnumbered]{algorithm2e}
\usepackage{float}

\usepackage{graphicx}
\usepackage{booktabs}
\usepackage{array}
\usepackage{tabularx}
\usepackage{makecell}

\usepackage{cite}
\usepackage{url}
\usepackage{xcolor}
\definecolor{linkblue}{HTML}{1F4E79}
\usepackage[colorlinks=true,linkcolor=black,
  citecolor=green,urlcolor=linkblue]{hyperref}
\usepackage{balance}

\newcommand{\ActAccept}{\ensuremath{\mathsf{Accept}}\xspace}
\newcommand{\ActInspect}{\ensuremath{\mathsf{Inspect}}\xspace}
\newcommand{\ActRelabel}{\ensuremath{\mathsf{Relabel}}\xspace}
\newcommand{\ActArbitrate}{\ensuremath{\mathsf{Arbitrate}}\xspace}
\newcommand{\ActUnresolved}{\ensuremath{\mathsf{Unresolved}}\xspace}
\newcommand{\ActA}{\mathsf{A}}
\newcommand{\ActI}{\mathsf{I}}
\newcommand{\ActR}{\mathsf{R}}
\newcommand{\ActD}{\mathsf{D}}
\newcommand{\ActU}{\mathsf{U}}

\newsavebox{\JABERDropCapBox}
\newlength{\JABERDropCapWidth}
\renewcommand{\IEEEPARstart}[2]{%
  \par\noindent
  \sbox{\JABERDropCapBox}{{\fontsize{24}{24}\selectfont\bfseries #1}}%
  \setlength{\JABERDropCapWidth}{\wd\JABERDropCapBox}%
  \addtolength{\JABERDropCapWidth}{0.15em}%
  \hangindent=\JABERDropCapWidth
  \hangafter=-2
  \makebox[0pt][l]{%
    \hspace{-\JABERDropCapWidth}%
    \raisebox{-0.88\baselineskip}[0pt][0pt]{\usebox{\JABERDropCapBox}}}%
  {\MakeUppercase{#2}}%
}

\newtheoremstyle{jaberplain}
  {\topsep}
  {\topsep}
  {\itshape}
  {}
  {\bfseries}
  {.}
  {5pt plus 1pt minus 1pt}
  {\thmname{#1}\thmnumber{ #2}\thmnote{\normalfont\itshape\ (#3)}}
\newtheoremstyle{jaberdefinition}
  {\topsep}
  {\topsep}
  {\normalfont}
  {}
  {\bfseries}
  {.}
  {5pt plus 1pt minus 1pt}
  {\thmname{#1}\thmnumber{ #2}\thmnote{\normalfont\itshape\ (#3)}}
\theoremstyle{jaberplain}
\newtheorem{assumption}{Assumption}
\newtheorem{proposition}{Proposition}

\newtheorem{lemma}{Lemma}
\theoremstyle{jaberdefinition}
\newtheorem{definition}{Definition}
\theoremstyle{remark}
\newtheorem{remark}{Remark}
\usepackage{microtype}
\DeclareMicrotypeAlias{texgyrepagella-regular.otf}{Palatino}
\usepackage[table]{xcolor}

\usepackage{titlesec}

\titlespacing*{\section}
{0pt}    
{3pt}    
{2pt}    

\titlespacing*{\subsection}
{0pt}    
{3pt}    
{2pt}    

\begin{document}

\title{\vspace{-5mm} JABER: Sequential Quality Control for Human--AI Data Annotation under Shared Bias \vspace{-2mm}}

\author{Fengxu Yang, Minghui Liwang,~\IEEEmembership{Senior Member,~IEEE,}
Yang Lu,~\IEEEmembership{Senior Member,~IEEE,} Bo Han,\\
Xianbin Wang,~\IEEEmembership{Fellow,~IEEE}
and Seyyedali Hosseinalipour,~\IEEEmembership{Senior Member,~IEEE}%
\thanks{F. Yang (2451368@tongji.edu.cn) is with the Guohao School, Tongji University, Shanghai, China. Minghui Liwang (minghuiliwang@tongji.edu.cn) is with the Department of
Control Science and Engineering and the National Key Laboratory of Autonomous
Intelligent Unmanned Systems, Tongji University,
Shanghai, China. Y. Lu (luyang@xmu.edu.cn) is with the School
of Informatics, Xiamen University, Fujian, China. B. Han (bhanml@comp.hkbu.edu.hk) is with Department of Computer Science, Hong Kong Baptist University, Hong Kong, China. Xianbin Wang (xianbin.wang@uwo.ca) is with the Department of Electrical and Computer Engineering,
Western University, Ontario, Canada. Seyyedali Hosseinalipour (alipour@buffalo.edu) is
with the Department of Electrical Engineering, University at Buffalo-SUNY,
Buffalo, NY USA.}}


\IEEEtitleabstractindextext{%
\begin{abstract}
\setlength{\rightskip}{0pt}
High-quality annotations are essential for training and evaluating artificial intelligence (AI)
systems. To ensure annotation reliability, quality control (QC) mechanisms often acquire additional
human or AI judgments before accepting a label. Nevertheless, shared biases among annotators can
lead to correlated judgments, causing conventional methods to mistake repeated agreement for
independent evidence of correctness. To address this challenge, we propose JABER (Joint
Alias-Preserving Belief for Evidence Routing), a bias-aware sequential QC framework that maintains a
joint Bayesian posterior over label correctness and shared biases to guide evidence
acquisition and termination decisions. Building on this posterior, we develop a finite-horizon
decision-making algorithm that selects evidence-acquisition and termination actions to minimize
the expected QC loss under resource constraints. Our theoretical analysis demonstrates that conventional
methods that update label-correctness probabilities under the assumption of conditionally
independent annotator judgments can become arbitrarily confident in an incorrect label under
persistent shared bias. We further identify conditions under which repeated inspections cannot
distinguish a correct label from an incorrect label supported by shared bias, establish when
obtaining a new label can resolve this ambiguity, and demonstrate why modeling
the dependence between label correctness and shared bias can be necessary for QC decisions.
Our experiments show that JABER reduces expected QC loss
by an average of 4.65\% compared with a method that maintains separate label correctness and bias
probabilities under conditions where shared bias makes repeated inspections unreliable. Further,
JABER's finite-horizon planning, which evaluates the potential outcomes of multiple successive QC
actions, reduces expected QC loss by an average of 10.05\% compared with
a one-step lookahead method. In real-application programming interface (real-API) experiments on FEVER, JABER achieves zero labeling errors while reducing QC costs by 2.67\% and 16.99\% compared
with marginal-based belief modeling and one-step lookahead methods, respectively.
\end{abstract}
\vspace{-2mm}
\begin{IEEEkeywords}
Human--AI collaborative annotation, sequential quality control, shared latent bias, finite-horizon belief planning.
\end{IEEEkeywords}
}
\maketitle
\IEEEdisplaynontitleabstractindextext

\setlength{\abovedisplayskip}{2pt}
\setlength{\belowdisplayskip}{2pt}

\IEEEraisesectionheading{\section{Introduction}}
\IEEEPARstart{T}{he} increasing deployment of artificial intelligence (AI) in high-stakes
applications has made reliable data annotation essential. Modern annotation pipelines increasingly
combine human annotators, large language model (LLM) labelers, and LLM-as-a-judge systems for
dataset construction, preference assessment, and model
evaluation~\cite{panickssery2024llm,wang2024llm,liu2025human}. However, such human--AI collaboration
can introduce shared biases that lead to correlated judgments, making it difficult for quality
control (QC) mechanisms to distinguish independent evidence of label correctness from agreement
driven by common biases. This challenge is particularly pronounced in LLM-assisted annotation, as
evaluators may exhibit correlated judgments due to shared pretraining data, model families, prompts,
or evaluation criteria. Similarly, human annotators may exhibit correlated judgments when exposed to
common interfaces and task contexts. Such dependencies can be further amplified by task-level cues,
including response length, presentation order, and model provenance. Consequently, high agreement
among annotators may reflect shared biases rather than independent evidence of label
correctness~\cite{care2026confounder,ising2026aggregation,kelly2025correlated}, effectively reducing
the number of independent judgments~\cite{kim2025correlated,kohli2026nine}.

Moreover, human and AI errors are not necessarily complementary~\cite{liu2025human}, making
correlated judgments particularly problematic in sequential QC. For instance, when multiple
annotators agree on a label because of a shared bias, conventional methods that assume independent
judgments may mistakenly interpret this agreement as strong evidence of correctness. As a result,
the system may become overly confident and accept an incorrect label. Although existing
dependence-aware methods account for such correlations when combining a fixed set of
judgments~\cite{ising2026aggregation,care2026confounder}, post-annotation QC introduces an
additional challenge: deciding what to do after each new judgment. Specifically, the system must
determine whether to accept the current label, request another inspection, obtain a replacement
label, seek arbitration, or leave the task unresolved. Since these actions differ in cost,
reliability, and sensitivity to shared bias, choosing the best action requires considering both the
evidence already collected and the potential value of acquiring further evidence. Thus, accounting
for shared bias is essential not only for accurately assessing label correctness but also for
determining when and how to continue QC.

\subsection{Core Motivation}
In this work, we investigate the following three research questions (RQs) that motivate our
research:

\noindent\textbullet\ \textit{RQ1: How should evidence from human and AI annotators be evaluated
during QC when agreement on a label may result from shared bias rather than independent support for
its correctness?}
Existing methods account for correlations among human and AI judgments when combining multiple
annotations to estimate label correctness~\cite{ising2026aggregation,care2026confounder}. However,
repeated agreement can reflect either label correctness or a persistent bias favoring an incorrect
label. When a QC action produces identical observation probabilities under these two explanations,
the action cannot distinguish between them, a phenomenon known as \textit{aliasing}.

\noindent\textbullet\ \textit{RQ2: When should sequential annotation QC stop collecting additional
judgments and instead use a more informative evidence source to resolve uncertainty about label
correctness?}
Collecting additional judgments is useful only when they provide new information about whether the
current label is correct. However, under persistent shared bias, repeated reviews of the same label
may produce similar judgments regardless of its actual correctness, making further inspections
ineffective at resolving uncertainty. In such cases, the system may benefit from switching to a
different QC action.

\noindent\textbullet\ \textit{RQ3: How can belief, cost, and future decision opportunities be
integrated in sequential QC?}
Choosing the best QC action requires considering not only its immediate cost and outcome but also
how it may improve subsequent decisions. The remaining budget and decision steps also constrain
which follow-up actions remain available.

\subsection{Novelty and Contribution}
To address the above RQs, we propose \textit{Joint Alias-Preserving Belief for Evidence Routing}
(\textit{JABER}), a framework for sequential QC in human--AI collaborative annotation. Our main
contributions can be summarized as follows:

\noindent\textbullet\ \textit{An alias-preserving joint belief model for correlated judgments.} We
develop a bias-aware Bayesian framework that jointly models label correctness and persistent shared
biases through an action-dependent observation model. Unlike conventional approaches that may
mistake repeated agreement for independent evidence of correctness, our framework maintains a joint
posterior that preserves competing explanations for observed judgments. Furthermore, we derive
Bayesian update rules that account for newly acquired evidence and changes in the held label.

\noindent\textbullet\ \textit{A theoretical characterization of evidence distinguishability and the
need for joint beliefs.} We establish theoretical results revealing how persistent shared biases can
cause conventional conditional-independence label updating methods to become increasingly confident
in an incorrect label, while repeated inspections may fail to distinguish genuine correctness from
bias-induced agreement. Building on these findings, we derive conditions under which the \ActRelabel
action, which obtains a new label rather than merely assessing the existing one, provides additional
information to distinguish these competing explanations. Furthermore, we demonstrate that two joint
beliefs with identical correctness and bias marginals can lead to different optimal QC actions,
establishing the importance of preserving their joint dependence for informed decision-making.

\noindent\textbullet\ \textit{A finite-horizon planning algorithm for sequential quality control.}
We develop a Bellman recursion for a finite-horizon partially observable Markov decision process
(POMDP)~\cite{dai2013pomdp} to minimize the expected total QC loss. Unlike approaches that consider
only immediate outcomes, our algorithm accounts for the costs of each action, the information gained
from new observations, potential label replacements, and the decisions available in subsequent
steps. By incorporating the remaining budget and decision steps, our algorithm determines
the optimal QC action at each stage and adaptively replans as new observations become available. We
establish that this procedure yields the optimal policy under the specified observation model and
known parameters.

\noindent\textbullet\ \textit{Empirical evaluation and applicability.} We conduct controlled synthetic experiments, including ablation studies and robustness analyses, to quantify
the benefits of joint belief modeling and finite-horizon planning in reducing QC loss. Furthermore,
we implement a controlled real-API QC workflow on FEVER~\cite{thorne2018fever}, and analyze judgment
records from HealthContradict~\cite{zhang2026healthcontradict} and four additional public datasets
to investigate correlated annotation errors and identify how evidence informativeness, the
reliability of replacement judgments, and action costs affect the value of further QC.
\section{Literature Review}
Henceforth, we review three related research directions: sequential QC in annotation
systems, dependence-aware modeling of correlated human--AI judgments, and human--AI routing and
deferral. For each direction, we clarify how its objective and decision setting differ from those of
JABER.

\subsection{Sequential Quality Control in Annotation Systems}
Classical annotation QC focuses on recovering reliable labels from noisy judgments. In this context,
the Dawid--Skene model jointly estimates latent labels and annotator-specific error
rates~\cite{dawid1979maximum}, while later studies show that annotator reliability may vary
substantially across tasks and instances~\cite{burrell2023testing}. Related works have
further modeled dependencies among noisy labeling
sources~\cite{ratner2017snorkel,varma2019learning}, allocated limited review budgets to samples
selected for re-annotation~\cite{bernhardt2022active}, and studied iterative correction when
relabeling itself remains imperfect~\cite{peng2026relabeling}. Beyond improving label estimation,
sequential QC introduces the challenge of deciding which additional judgments to acquire and when to
terminate the review process. To address this challenge, POMDP-based workflow control balances
output quality against evidence acquisition cost~\cite{dai2013pomdp}, while MDP-based sequential
crowdsourcing incorporates accuracy, cost, and time into annotation
decisions~\cite{raykar2014sequential}. Other approaches focus on specific aspects of adaptive QC.
For example, CLARA adaptively determines review counts from evolving
confidence~\cite{nguyen2020clara}, partitioning Monte Carlo methods support time-constrained
decisions~\cite{deng2025partitioning}, and Bayesian online learning for consensus prediction (BOCP)
predicts the label favored by the full expert panel using partial
feedback~\cite{showalter2024consensus}. Similarly, collaborative adaptive labeling (CAL) combines
repeated low-cost annotations with selective escalation to expert reviewers~\cite{ruan2026cal}.

Although these methods demonstrate the benefits of adaptively acquiring additional judgments and
deciding when to stop QC, they do not explicitly maintain a joint belief over label correctness and
persistent shared bias for sequential QC. In particular, repeated support for a label may indicate
either that the label is correct or that a shared bias consistently favors an incorrect label. This
limitation is especially relevant to CAL~\cite{ruan2026cal}, which considers correlated annotations
in its experiments but does not track that bias in its QC belief updates. To address this gap, JABER
maintains a joint posterior over current-label correctness and persistent shared bias, preserving
these competing explanations and using the resulting belief to guide subsequent QC actions and
termination decisions.

\subsection{Correlated Errors and Shared Bias in Human--AI Annotation}
LLM-based judges are increasingly used for data filtering and automated
annotation, yet their errors may exhibit systematic dependence. For instance, self-preference,
response length, presentation order, and evaluation protocols can consistently influence their
assessments~\cite{panickssery2024llm,wang2024llm,saito2023verbosity,li2024split}. These dependencies
can be further amplified when judges share model families, training data, prompts, or evaluation
criteria, leading to similar errors across different judges. Indeed, empirical studies have reported
substantial error correlations among LLM judges~\cite{kim2025correlated}, suggesting that the amount
of independent evidence provided by multiple judgments may be considerably smaller than the number
of judges involved~\cite{kohli2026nine}. Motivated by these findings, recent studies have developed
methods to explicitly account for such dependencies. For example, Kelly et
al.~\cite{kelly2025correlated} learn a joint latent representation of human experts and classifiers
to predict missing expert judgments and reduce the need for additional queries. Other approaches
focus on modeling correlations when combining judgments from multiple evaluators. Specifically,
Ising-based aggregation captures pairwise dependencies among judges~\cite{ising2026aggregation},
while CARE~\cite{care2026confounder} distinguishes underlying label quality from shared factors that
may systematically influence multiple LLM judges.

JABER differs from the above approaches in how dependence guides QC of an existing label: rather
than accounting for correlations primarily to improve label aggregation or predict missing expert
judgments, JABER jointly models the correctness of the current label and the presence of persistent
shared biases. By maintaining and updating this joint belief as new judgments are collected or the
current label is replaced, JABER preserves competing explanations for observed agreement, whether it
reflects genuine label correctness or shared bias. More importantly, this joint belief guides the
selection of subsequent QC actions, enabling the system to determine whether additional evidence is
worthwhile and when to terminate the review process.

\subsection{Human--AI Routing and Learning to Defer}
Learning to Defer (L2D) studies when an automated system should make a prediction independently and
when it should rely on a human expert~\cite{mozannar2020consistent}. Building on this idea,
subsequent studies have considered dependencies between human and AI prediction
errors~\cite{wei2024exploiting} and developed methods for assigning tasks to experts with different
capabilities~\cite{mao2025mastering}. Similarly, model-routing approaches select among AI models or
services based on their prediction quality and cost~\cite{ding2024hybrid}. In the context
of data annotation, QUORUM~\cite{purificato2026quorum} optimizes the allocation of a fixed
annotation budget across human and LLM annotators while allowing multiple judgments for the same
item. From a complementary perspective, active sequential hypothesis testing provides mathematical
foundations for selecting observations that can effectively distinguish among competing
hypotheses~\cite{naghshvar2013active}.

Although these approaches provide valuable insights into assigning tasks to different annotators and
selecting informative observations, they do not directly address the problem of sequentially
verifying an existing label under persistent shared bias. In contrast, JABER starts with an already
assigned label and determines which QC action should be performed next to minimize the expected QC
loss. Specifically, it considers whether to accept the current label (\ActAccept), request another
judgment to assess its correctness (\ActInspect), obtain a new label that may replace the existing
one (\ActRelabel), refer the task for a final decision (\ActArbitrate), or terminate without
assigning a final label (\ActUnresolved). To make these decisions, JABER combines Bayesian inference
that accounts for correlated judgments with finite-horizon planning, allowing it to evaluate the
potential benefits of future actions while accounting for their costs, the risk of accepting an
incorrect label, and the available resources.
\section{Core Modeling}
\label{sec:overview}

\subsection{Motivating Example and Main Idea}
\label{subsec:toy-example}

Consider a pairwise preference annotation task in reinforcement learning from human feedback (RLHF)
or LLM-as-a-judge evaluation. Given a prompt and two candidate responses, $u$ and $v$, the goal is
to determine which response is preferred by assigning a label from the binary set
$\mathcal{Y}=\{u\succ v,\ v\succ u\}$, where $u\succ v$ indicates that response $u$ is preferred to
$v$, and vice versa. Let $\mathbf{x}$ denote the observable task attributes at the start of
sequential QC, including the initial annotation source and other bias cues such as response length,
position, and model family. Suppose that a human or LLM annotator produces the initial preference
label $\hat{y}_{\mathrm{init}}=u\succ v$. The preferred response $u$ may be longer than $v$,
presented first, or generated by a model from the same family as a subsequent judge. Although these
cues do not establish that $u$ is superior to $v$, they may induce correlated preferences across
judges and inflate the apparent strength of agreement. To assess the correctness of the initial
label, the platform then applies \ActInspect, which asks another human or LLM judge whether the held
label is supported. This operation returns pass or fail and updates the belief without replacing the
held label. Suppose that the first \ActInspect returns pass, indicating support for $u\succ v$. If
the platform continues to use judge models exposed to similar biases, a second \ActInspect may also
return pass. Under conditional-independence updating, these two pass outcomes are treated as
independent evidence, successively increasing confidence that the held label is correct. As a
result, the platform may choose \ActAccept, thereby accepting $u\succ v$ and terminating the task.
\textit{However, the two pass outcomes may reflect the same verbosity, position, or self-preference
bias rather than independent confirmation of label correctness.} Consequently, the platform may
become overconfident and prematurely accept an incorrect initial label.

\begin{figure}[t]
\vspace{-3mm}
\centering
\includegraphics[width=0.49\textwidth,trim=30bp 14bp 18bp 5bp,clip]{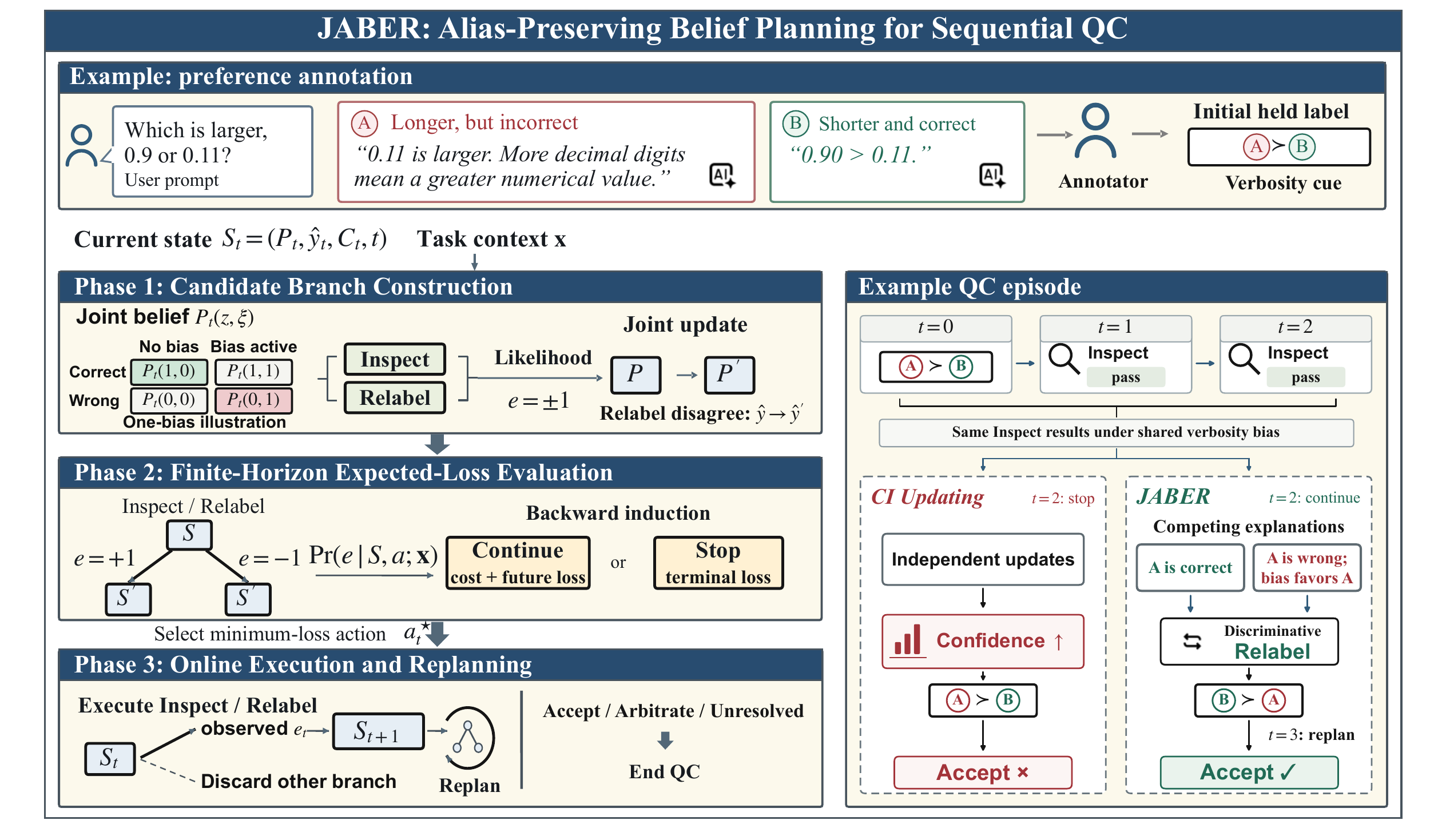}
\vspace{-8mm}
\caption{
Overview of JABER framework for sequential annotation QC under shared bias. The top panel introduces a motivating preference-annotation example in which an annotator initially favors a longer but incorrect response, potentially due to verbosity bias. The left panel illustrates JABER’s three phases: constructing possible observation outcomes and updating the joint belief over label correctness and shared bias (Phase 1), evaluating the expected losses of feasible QC actions through finite-horizon planning (Phase 2), and executing the selected action followed by belief updating and replanning (Phase 3). The right panel shows how repeated agreement driven by shared bias can lead conventional conditional-independence updating to accept the incorrect label, whereas JABER preserves competing explanations and uses a more informative QC action to identify the correct label.}
\label{fig:jaber-framework}
\vspace{-5mm}
\end{figure}

Motivated by the above example, we design JABER, which jointly models current-label correctness and
the shared-bias state. Specifically, the same two pass outcomes can have two competing explanations:
the held label may be correct, or it may be incorrect but repeatedly supported by judges due to a
persistent shared bias. To account for this ambiguity, JABER maintains a joint posterior that
preserves both explanations. Fig.~\ref{fig:jaber-framework} illustrates the motivating example and
the overall framework of JABER. In the following, we formalize the sequential QC task, latent
states, joint belief, shared-bias observation model, and exact belief updates.

\subsection{Sequential QC Setting and Actions}
\label{subsec:system-overview}
\label{subsec:task-init}
\label{subsec:action-space}

We use a binary label space $\mathcal{Y}$ containing two mutually exclusive values. Both the true
label $y^{*}$ and initial label $\hat{y}_{\mathrm{init}}$ belong to $\mathcal{Y}$, where
$|\mathcal{Y}|=2$. Also, the held label, meaning the platform's current working label, is
initialized as $\hat{y}_0=\hat{y}_{\mathrm{init}}$. For each task, the platform initiates a QC
episode after the initial label is generated. It can collect successive judgments without revealing earlier ones to reviewers. Each episode may involve multiple decision steps
indexed by $t=0,1,\ldots$, where the held label at step $t$ is denoted by $\hat{y}_t\in\mathcal{Y}$.

We consider four QC actions, forming $\mathcal A=\{\ActA,\ActI,\ActR,\ActD\}$. Among these,
\ActInspect ($\ActI$) and \ActRelabel ($\ActR$) are continuing actions: \ActInspect asks another
human or AI judge to review $\hat{y}_t$, using the resulting pass/fail observation to update the
joint posterior without changing the held label, while \ActRelabel obtains a new label, compares it
with $\hat{y}_t$, updates the joint posterior, and replaces the held label if the two disagree.
After either of the above actions, the episode proceeds to the next QC step. On the other hand,
\ActAccept ($\ActA$) and \ActArbitrate ($\ActD$) are terminal actions: \ActAccept outputs the
current held label $\hat{y}_t$ and terminates the QC episode without producing a new observation or
updating the posterior, while \ActArbitrate escalates the task for a final judgment and terminates
the episode with the arbitrated label. Since no further decisions are made, the arbitration result
is not used to update the joint posterior. Finally, we define \ActUnresolved ($\ActU$) as an
additional terminal outcome, which ends the episode without acquiring new evidence or producing a
final label, returning $\varnothing$. Accordingly, the extended decision set is
$\widetilde{\mathcal A}=\mathcal A\cup\{\ActU\}$.

\subsection{Hidden States and Joint Belief}
\label{subsec:hidden-state}

To consider correlations among judgments during sequential QC, we introduce two latent states
representing the correctness of the current label and the presence of shared biases. We
then define a joint belief over these states, allowing the platform to update its understanding of
label correctness and shared bias as new observations become available.
\begin{definition}[Latent Correctness and Shared-Bias State]
\label{def:latent-state}
Define the current-label correctness indicator as
\begin{equation}
z_t = \mathbf{1}[\hat{y}_t = y^{*}] \in \{0, 1\},
\label{eq:correctness-state}
\end{equation}
which specifies whether the held label at step $t$ matches the true label.
We further define the shared-bias state as
\begin{equation}
\xi = (\xi_1, \xi_2, \ldots, \xi_K) \in \{0, 1\}^K,
\label{eq:shared-bias-state}
\end{equation}
where $\xi_k=1$ indicates that bias type $k$ is active. Shared bias is modeled as an episode-level
latent factor that, when active, can induce systematic shifts in judgments across
multiple QC steps.
The finite set of $K$ bias types captures different sources of bias in human judgments
and in LLMs.
\end{definition}

\begin{assumption}[Episode-Level Persistence of Shared Bias]
\label{asmp:static-xi}
Correlated annotation errors may arise when multiple human or AI annotators share common sources of
bias, such as model provenance or evaluation
protocols~\cite{panickssery2024llm,wang2024llm,kim2025correlated}. To capture the persistent
influence of such biases across successive QC steps, we assume that the shared-bias state $\xi$
remains fixed throughout each QC episode, although it may vary across different episodes.
\end{assumption}

Building on these latent-state definitions, we next introduce the joint posterior and its marginal
probabilities. Given the observable task attributes $\mathbf{x}$, such as judgment source and
response characteristics, the platform maintains a joint belief over the latent state $(z_t,\xi)$,
as characterized below.
\begin{definition}[Joint Posterior and Marginal Readouts]
\label{def:joint-posterior}
Before the decision at step $t$, the joint posterior table $P_t$ represents the probability
distribution over the latent state $(z_t,\xi)$ after incorporating all preceding QC observations.
Accordingly, letting $z$ denote a possible value of $z_t$, we have
\begin{equation}
\begin{gathered}
P_t(z,\xi)\geq 0,\qquad
z\in\{0,1\},\quad \xi\in\{0,1\}^{K},\\
\sum_{z\in\{0,1\}}\sum_{\xi\in\{0,1\}^K}
P_t(z,\xi)=1 .
\end{gathered}
\label{eq:joint-belief}
\end{equation}
The label $\hat y_t$ and task attributes $\mathbf{x}$ determine the observation likelihood in
the belief update, but are not additional arguments of $P_t$ (the update rule for $t>0$ is given in
Section~\ref{subsec:belief-update}).
From this joint posterior, we compute the probability $b_t$ that the current label is correct and
the bias-activity belief vector $\bm{\rho}_t=(\rho_{t,1},\ldots,\rho_{t,K})$, whose $k$-th component
represents the probability that bias type $k$ is active, by marginalization (i.e., by summing over
the remaining latent variables):
\begin{equation}
\begin{aligned}
b_t&=\sum_{\xi\in\{0,1\}^K}P_t(1,\xi)\in[0,1],\\
\rho_{t,k}&=\sum_{z\in\{0,1\}}\sum_{\xi:\xi_k=1}P_t(z,\xi)\in[0,1].
\end{aligned}
\label{eq:joint-marginals}
\end{equation}
Also, the log-odds of label correctness is
$\lambda_t{=}\log\!\bigl(b_t/(1-b_t)\bigr)$.
\end{definition}
To initialize the joint posterior, we use the following factorized distribution, where
$\mathrm{Bern}(u;p)=p^u(1-p)^{1-u}$:
\begin{equation}
P_0(z,\xi)=\mathrm{Bern}(z;b_0)
\prod_{k=1}^{K}\mathrm{Bern}(\xi_k;\rho_{0,k}).
\label{eq:joint-prior}
\end{equation}
This expression is used only at $t=0$, where label
correctness and the bias states are assumed to be independent. After an \ActInspect or
\ActRelabel action, the new observation updates the joint posterior, potentially introducing
dependencies between label correctness and shared bias. The initial probabilities $b_0$ and
$\bm\rho_0$ are specified/calibrated for the task.

\subsection{Shared-Bias Observation Model}
\label{subsec:observation-model}
\label{subsec:action-observation-interface}

For \ActInspect, \ActRelabel, and \ActArbitrate, which produce binary outcomes, we encode the
observation at step $t$ as $e_t\in\{+1,-1\}$, with its interpretation depending on the selected
action $a_t$. Specifically, under \ActInspect, $e_t=+1$ and $e_t=-1$ indicate pass and fail,
respectively, that is, support for and opposition to the held label $\hat{y}_t$. Under \ActRelabel
and \ActArbitrate, $e_t=+1$ indicates that the newly obtained label agrees with $\hat{y}_t$, whereas
$e_t=-1$ indicates disagreement.

\begin{definition}[Bias-Cue Direction and Intensity]
\label{def:cue-intensity}
A bias cue refers to an observable task characteristic that may systematically favor or oppose a
particular label when the corresponding shared bias is active, but its presence alone does not
establish whether that bias is active. For shared bias type $k$, let $s_{t,k}\in\{-1,0,+1\}$ denote
the cue direction, where $+1$ denotes a cue favoring the held label, $-1$ indicates a cue opposing
it, and $0$ indicates the absence of a cue. Further, let $m_{t,k}\in[0,1]$ denote
the cue intensity determined by the task attributes. For example, if the held label is $u\succ v$
and response $u$ is longer than $v$, the cue associated with verbosity bias (VB) may take
$s_{t,k}=+1$, indicating a tendency to favor the held label. Combining the cue direction and
intensity, we define the directional strength as
\begin{equation}
r_{t,k} = s_{t,k} \cdot m_{t,k}, \qquad r_{t,k} \in [-1, +1].
\label{eq:r-def}
\end{equation}
In essence, $\mathbf{x}$ specifies $m_{t,k}$ and, together with $\hat{y}_t$, $s_{t,k}$;
\eqref{eq:r-def} then gives $r_{t,k}$. The subscript $t$ emphasizes that the cue direction depends
on $\hat{y}_t$, meaning that $s_{t,k}$ and $r_{t,k}$ may reverse when the \ActRelabel action
replaces the current label.
\end{definition}

\begin{definition}[Action Diagnostic Strength and Bias Coupling]
\label{def:action-coupling}
Let $w_a > 0$ denote the baseline diagnostic strength of action $a$ in the absence of shared-bias
effects, with larger values indicating more reliable judgments. We assume
\begin{equation}
w_{\ActD} > w_{\ActR} \geq w_{\ActI} > 0.
\label{eq:w-order}
\end{equation}
Furthermore, let $\varphi_{a,k}\geq 0$ represent the coupling strength between action $a$ and bias
type $k$, thereby controlling the effect of the directional cue $r_{t,k}$ on the observations
produced by action $a$. Accordingly, we define the total bias-induced shift for action $a$ under the
shared-bias state $\xi$ as
\begin{equation}
\Delta_t(a, \xi) \;=\; \sum_{k=1}^{K} \varphi_{a,k} \, \xi_k \, r_{t,k}.
\label{eq:Delta-def}
\end{equation}
The aggregate shift $\Delta_t(a,\xi)$ and baseline diagnostic strength $w_a$ are measured on the
same scale and added in the observation model. The coupling coefficients collectively form the
matrix $\boldsymbol{\Phi}=[\varphi_{a,k}]$, allowing the same shared bias to affect \ActInspect,
\ActRelabel, and \ActArbitrate to different degrees. The cue-intensity calculation and scale are
fixed before the coupling coefficients are specified.
\end{definition}

Conditioned on the latent state and the QC action performed at the current step, we define the
following probabilistic observation model for $e_t$:
\begin{equation}
\hspace{-3.5mm}\resizebox{0.92\columnwidth}{!}{$
\Pr(e_t \mid z_t, \xi, a_t, \mathbf{x},\hat y_t) = \sigma\!\Bigl(
 e_t \cdot \bigl[(2z_t-1)w_{a_t}+\Delta_t(a_t,\xi)\bigr]
\Bigr),
$}\hspace{-3mm}
\label{eq:unified-obs}
\end{equation}
where $\sigma(\cdot)$ is the sigmoid function. In~\eqref{eq:unified-obs}, the baseline term
$(2z_t-1)w_{a_t}$ provides an unbiased discriminative signal about whether the label is correct,
while the shift term $\Delta_t(a_t,\xi)$ captures the systematic effects of shared bias.
Consequently, an observation that supports the held label does not necessarily imply that the label
is correct, as such support may also result from shared bias.

At step $t$, let $\mathbf r_t=(r_{t,1},\ldots,r_{t,K})$ denote the vector of directional cue
strengths associated with all bias types. Although individual observations are generated
probabilistically according to~\eqref{eq:unified-obs}, they may exhibit statistical dependence due
to their common underlying shared-bias state. To capture this dependence while keeping the
observation model tractable, we adopt the following conditional-independence assumption, which is
consistent with latent-factor models where observations are conditionally independent given the
underlying shared latent factors~\cite{ising2026aggregation}.

\begin{assumption}[Conditional Independence under the Modeled Latent State]
\label{asmp:cond-indep-z-xi}
Under the specified observation model, we assume 
\begin{equation}
\hspace{-3.5mm}\resizebox{0.92\columnwidth}{!}{$\Pr(e_t\mid e_0,\ldots,e_{t-1},z_t,\xi,a_t,\mathbf{x},\hat y_t)=\Pr(e_t\mid z_t,\xi,a_t,\mathbf{x},\hat y_t).
$}\hspace{-3.5mm}
\label{eq:iid-z-xi}
\end{equation}
In other words, given the current correctness state, shared-bias state, selected action, and task
context, previous observations provide no additional information about $e_t$. Importantly, this
assumption does not imply that observations are independent when the shared-bias state is unknown.
Rather, it implies that observations are conditionally independent given the modeled latent states,
while they may remain statistically dependent after marginalizing over these states due to their
shared underlying bias. This distinction is essential because it allows our model to capture
correlations among successive judgments induced by persistent shared biases, rather than incorrectly
treating repeated agreement as independent evidence of label correctness.
\end{assumption}
Having established the observation model and its conditional-independence assumption, we next derive
the exact joint posterior updates for \ActInspect and \ActRelabel.

\subsection{Exact Joint Update and Label Switching}
\label{subsec:belief-update}

The exact joint posterior updates described below apply only to $a_t\in\{\ActI,\ActR\}$, since the
remaining actions, \ActAccept, \ActArbitrate, and \ActUnresolved, are terminal actions that end the
QC episode without belief updates. Within our finite-state model, updating the joint
posterior involves enumerating all states represented by $P_t(z,\xi)$. Let
$\mathcal{L}_t(z,\xi)=\Pr(e_t\mid z,\xi,a_t,\mathbf{x},\hat y_t)$ denote the likelihood of the
current observation, as specified by~\eqref{eq:unified-obs}. Since \ActRelabel may replace the held
label, it can also change the interpretation of the correctness state $z$. To consistently handle
both \ActInspect and \ActRelabel, we first incorporate the observation while keeping the held label
$\hat y_t$ and the current indexing of $z$ unchanged. Specifically, we perform the following
full-state Bayesian update:
\begin{equation}
\hspace{-3mm}\widetilde{P}_{t+1}(z,\xi)
=
\frac{P_t(z,\xi)\mathcal{L}_t(z,\xi)}
{\sum_{z'\in\{0,1\}}\sum_{\xi'\in\{0,1\}^K}
P_t(z',\xi')\mathcal{L}_t(z',\xi')}.
\hspace{-3mm}
\label{eq:joint-bayes-update}
\end{equation}
The resulting $\widetilde{P}_{t+1}$ is a temporary joint posterior that incorporates the
current observation but does not yet account for potential label switching. From this posterior, we
obtain
\[
\widetilde{b}_{t+1}=\hspace{-2mm}\sum_{\xi\in\{0,1\}^K}\widetilde{P}_{t+1}(1,\xi),
~~~~~
\widetilde{\lambda}_{t+1}=\log\!\left({\widetilde{b}_{t+1}}\big/{(1-\widetilde{b}_{t+1})}\right).
\]

For \ActInspect, we set $a_t=\ActI$ in $\mathcal L_t$ and leave $\hat y_t$ unchanged. Since no label
switching occurs, we have $z_{t+1}=z_t$, and the temporary posterior becomes the updated joint
posterior:
\[
\begin{aligned}
P_{t+1}(z,\xi)=\widetilde{P}_{t+1}(z,\xi),~~~~~~~~
\lambda_{t+1}=\widetilde{\lambda}_{t+1}.
\end{aligned}
\]

For \ActRelabel, we set $a_t=\ActR$ in $\mathcal L_t$ to obtain $\widetilde{P}_{t+1}$ and then use
the observation to determine whether the held label should be replaced: if the newly
obtained label agrees with the current label ($e_t=+1$), the held label and its correctness
indicator remain unchanged, giving $z_{t+1}=z_t$, whereas disagreement switches the held label and
gives $z_{t+1}=1-z_t$. Accordingly, the updated joint posterior is
\begin{equation}
P_{t+1}(z,\xi) \;=\;
\begin{cases}
\widetilde{P}_{t+1}(z,\xi),        & e_t = +1, \\[3pt]
\widetilde{P}_{t+1}(1-z,\xi),     & e_t = -1.
\end{cases}
\label{eq:relabel-update}
\end{equation}
When $e_t=-1$, the transformation $z\leftrightarrow1-z$ also gives $\lambda_{t+1}=-\widetilde{\lambda}_{t+1}$.
The updated beliefs $b_{t+1}$ and $\bm{\rho}_{t+1}$ are then obtained from $P_{t+1}$ using~\eqref{eq:joint-marginals}. Overall, the above updates allow JABER to consistently incorporate new observations while accounting for potential changes in the held label and preserving the shared-bias state.

\section{Problem Formulation}
\label{sec:problem-formulation}
\label{subsec:P0}

We formulate QC in human--AI collaborative annotation under persistent shared bias as a sequential decision-making problem, denoted by $\bm{\mathcal P}_0$, subject to a finite evidence-acquisition budget and a limit on the number of continuing QC steps. Our objective is to minimize the expected total QC loss over an episode by balancing cumulative action costs, terminal error risk, and the penalty for leaving a task \ActUnresolved{}. To this end, $\bm{\mathcal P}_0$ seeks a policy $\pi$ that maps the current belief state and remaining resources to the next feasible QC action at each step. In the following, we first define the information available to the policy and the feasible action set, followed by the termination conditions and finally the overall optimization formulation.

\smallskip\noindent\textbullet~\textit{Policy information and feasible actions.}
Recall the action set $\mathcal A=\{\ActA,\ActI,\ActR,\ActD\}$ and its extension $\widetilde{\mathcal A}=\mathcal A\cup\{\ActU\}$ (Section~\ref{subsec:action-space}). At each step, the policy $\pi$ selects an action $a_t$ based on the current joint posterior $P_t$, the label $\hat y_t$, the task attributes $\mathbf x$, the cumulative evidence-acquisition cost $C_t$, and the step index $t$. Here, $C_t$ accounts only for the costs of the \ActInspect and \ActRelabel actions already performed, with $C_0=0$.

For each $a\in\mathcal A$, let $c_a$ denote its execution cost, with $c_{\ActA}=0$. We assume that the evidence-acquisition budget $C_{\max}$ constrains only the costs of \ActInspect and \ActRelabel, while the \ActArbitrate cost $c_{\ActD}$ contributes to the total system cost but is excluded from the evidence-acquisition budget. Further, we let the positive integer $T_{\max}$ cap the number of continuing QC steps, and let $\mathcal A_{\mathrm{term}}=\{\ActA,\ActD,\ActU\}\subseteq\widetilde{\mathcal A}$ denote the terminal actions. Accordingly, the feasible action set at step $t$ is
\begin{equation}
\hspace{-3.5mm}\resizebox{0.92\columnwidth}{!}{$
\mathcal A_t^{\mathrm{feas}}=
\begin{cases}
\mathcal A_{\mathrm{term}},
& t\ge T_{\max},\\[0.35em]
\begin{aligned}
&\mathcal A_{\mathrm{term}}\cup\{a\in\{\ActI,\ActR\}:C_t+c_a\le C_{\max}\},
\end{aligned}
& t<T_{\max}.
\end{cases}
$}
\hspace{-3mm}
\label{eq:A-feas}
\end{equation}
As described in~\eqref{eq:A-feas}, all terminal actions remain available at every step, whereas \ActInspect and \ActRelabel can be selected if sufficient budget and decision steps remain. Once $t\ge T_{\max}$, no continuing action is allowed and the episode terminates. Thus, at each step, the policy selects a feasible action
\begin{equation}
a_t
=
\pi\bigl(P_t,\hat{y}_t,\mathbf{x},C_t,t\bigr)
\in
\mathcal A_t^{\mathrm{feas}}
\subseteq\widetilde{\mathcal A}.
\label{eq:policy}
\end{equation}
Finally, we let $\mathbb{P}^{\pi}$ denote the resulting probability distribution over latent states and action--observation trajectories.

\smallskip\noindent\textbullet~\textit{Stopping rule and terminal output.}
The QC episode terminates when the policy first selects a terminal action. Its stopping time is thus given by
\begin{equation}
T_{\mathrm{stop}}
=
\inf\bigl\{t: a_t \in \mathcal A_{\mathrm{term}}\bigr\},
\label{eq:Tstop}
\end{equation}
while the terminal action is $a_{T_{\mathrm{stop}}}\in\mathcal A_{\mathrm{term}}$. The final output $\hat{y}_{\mathrm{final}}$ is the held label $\hat y_{T_{\mathrm{stop}}}$ under \ActAccept, the arbitrated label $\hat y_{\mathrm{arb}}$ under \ActArbitrate, and $\varnothing$ under \ActUnresolved.

\smallskip\noindent\textbullet~\textit{Loss and optimization objective.}
An incorrect final output under \ActAccept or \ActArbitrate incurs an error cost $c_{\mathrm{err}}$, while terminating as \ActUnresolved incurs a penalty $c_{\ActU}$. Accordingly, we define the total loss over a QC episode as
\begin{equation}
\hspace{-3.5mm}\resizebox{0.92\columnwidth}{!}{$
\begin{aligned}
L_{\mathrm{episode}}
&={}\sum_{t=0}^{T_{\mathrm{stop}}-1}c_{a_t}
+c_{\ActD}\mathbf{1}[a_{T_{\mathrm{stop}}}=\ActD]\\
&+c_{\mathrm{err}}\mathbf{1}\!\left[
a_{T_{\mathrm{stop}}}\in\{\ActA,\ActD\},\
\hat y_{\mathrm{final}}\neq y^*
\right]+c_{\ActU}\mathbf{1}[a_{T_{\mathrm{stop}}}=\ActU].
\end{aligned}
$}\hspace{-4mm}
\label{eq:episode-loss}
\end{equation}
In~\eqref{eq:episode-loss}, the first term accumulates \ActInspect and \ActRelabel costs before termination; the second adds the arbitration cost when \ActArbitrate is selected; and the remaining terms penalize an incorrect final label or an unresolved outcome. Note that only the first term counts toward the evidence-acquisition budget $C_{\max}$, and the summation is zero if $T_{\mathrm{stop}}=0$.
Thus, for a given policy $\pi$, the expected episode loss is 
\begin{equation}
J(\pi)=\mathbb{E}_{\mathbb{P}^{\pi}}\!\left[L_{\mathrm{episode}}\right],
\label{eq:policy-expected-loss}
\end{equation}
where $\mathbb{E}_{\mathbb{P}^{\pi}}[\cdot]$ denotes the expectation over the latent-state and action--observation trajectory distribution $\mathbb{P}^{\pi}$. Our objective is therefore to find a policy that minimizes this expected loss while satisfying the budget and step constraints. Formally, the problem of our interest can be formulated as
\begin{equation}
\begin{aligned}
(\bm{\mathcal P}_0)\qquad
&\min_{\pi}\quad J(\pi)\\
&\text{subject to}\quad a_t\in\mathcal A_t^{\mathrm{feas}},
\quad 0\le t\le T_{\mathrm{stop}}.
\end{aligned}
\label{eq:P0}
\end{equation}
The feasibility constraint applies at every decision history reachable under $\pi$ and enforces the evidence-acquisition budget and the limit on continuing QC steps.

Solving $\bm{\mathcal P}_0$ requires finite-horizon sequential planning as continuing actions consume the available budget and decision steps. Specifically, performing \ActInspect or \ActRelabel incurs an immediate cost but may provide new information that improves subsequent QC decisions. Therefore, determining whether either action is worthwhile requires considering not only its immediate cost but also its potential to reduce future QC loss. A myopic approach that considers only immediate outcomes may overlook these benefits. We detail our proposed planning procedure in the next section and analyze its computational complexity in Appendix~\ref{app:implementation-details}.

\section{Methodology: JABER}
\label{sec:method}

In Section~\ref{sec:overview}, we established how new observations update the joint posterior over label correctness and shared bias. Building on this belief-update mechanism, we next develop JABER as a finite-horizon planning algorithm to solve $\bm{\mathcal P}_0$, which was formulated in Section~\ref{sec:problem-formulation}. In a nutshell, at each decision step, JABER constructs a belief tree rooted at the current state $S_t$ to evaluate the expected losses of feasible QC actions. For each continuing action (\ActInspect or \ActRelabel), the algorithm considers the possible observations, their probabilities, and the successor states. In contrast, terminal actions (\ActAccept, \ActArbitrate, and \ActUnresolved) are evaluated based on their immediate expected losses. Using backward induction from the tree leaves to the root, JABER accounts for the decisions available after each observation and selects the action $a_t^\star$ that minimizes the expected total remaining loss. Once the selected continuing action is executed, the algorithm retains only the successor state corresponding to the actual observation and replans from that state. JABER consists of three phases, where the evidence-acquisition budget and step constraints are enforced through the feasible action set at each decision step.

\subsection{Phase 1: Candidate Branch Construction}
\label{subsec:belief-engine}

In Phase~1, we construct the possible outcomes of each continuing QC action (\ActInspect and \ActRelabel) to support subsequent decision-making as follows.

\smallskip\noindent\textbullet~\textbf{Planning-state interface.}
At decision step $t$, the planner requires the current joint posterior, held label, cumulative evidence-acquisition cost, and current step index. We collect these quantities in a planning state
\begin{equation}
S_t=(P_t,\hat y_t,C_t,t).
\label{eq:planning-state}
\end{equation}
The latent state $(z_t,\xi)$ is unobserved; $P_t$ is the platform's belief over it, and $S_t$ is known to the planner. Here, $P_t$ denotes the joint posterior obtained by incorporating all preceding QC observations and accounting for label replacements. Furthermore, $\hat y_t$ is the held label, $C_t$ accumulates the costs of previously executed \ActInspect and \ActRelabel actions, and $t$ denotes the current decision step. Each node in the belief tree represents a planning state $S=(P,\hat y,C,t)$, with the root representing the current state $S_t$. The task-fixed attributes $\mathbf x$, which remain fixed throughout the QC episode, are passed separately as context. The pair $(S;\mathbf x)$ provides all the information needed to determine feasible actions, observation probabilities, successor states, and terminal losses without retaining the full action--observation history. Accordingly, the policy interface in~\eqref{eq:policy} can be written as $\pi(S;\mathbf x)$.

\smallskip\noindent\textbullet~\textbf{Candidate branch construction.}
Given a planning state $S$ and a continuing action $a\in\{\ActI,\ActR\}$, we consider the two possible observations $e\in\{+1,-1\}$. For each observation, we construct a candidate branch characterized by its occurrence probability and the successor state reached after the observation. Specifically, by marginalizing the observation model in~\eqref{eq:unified-obs} over the current joint posterior, we obtain the observation probability as
\begin{equation}
\hspace{-3.5mm}\resizebox{0.92\columnwidth}{!}{$
\begin{aligned}
\Pr(e\mid S,a;\mathbf x){=}\sum_{z\in\{0,1\}}\sum_{\xi\in\{0,1\}^K}
P(z,\xi)
\Pr(e\mid z,\xi,a,\mathbf x,\hat y).
\end{aligned}
$}
\hspace{-3.5mm}
\label{eq:predictive-e}
\end{equation}
Once the observation $e$ occurs, the corresponding successor state is determined by the transition
\begin{equation}
S'=\tau(S,a,e)=(P',\hat y',C+c_a,t+1),
\label{eq:state-transition}
\end{equation}
where $P'$ is the joint posterior after incorporating observation $e$, and $\hat y'$ is the held label after the action. In particular, if \ActRelabel returns disagreement $e=-1$, the held label is replaced and the joint posterior is transformed as in~\eqref{eq:relabel-update}.

Thus, each candidate branch is fully specified by its probability $\Pr(e\mid S,a;\mathbf x)$ and successor state $S'$. These branches serve as inputs to Phase~2, where the expected losses of \ActInspect and \ActRelabel are evaluated and compared against those of immediate termination.

\subsection{Phase 2: Finite-Horizon Expected-Loss Evaluation}
\label{subsec:evidence-routing}

Phase~2 evaluates the expected losses of feasible QC actions through backward induction to determine the optimal action at each decision step. Specifically, we first calculate the expected loss of terminating QC immediately (Step~1). We then recursively evaluate the expected losses of continuing actions over the remaining planning horizon (Step~2). Finally, we compare the resulting losses and select the action that minimizes the expected remaining loss (Step~3).

\smallskip\noindent\textbullet~\textbf{Step 1: Establishing the immediate-termination baseline.}
We first determine the expected loss of terminating QC at the current planning state. Recall that the three terminal actions are \ActAccept, \ActArbitrate, and \ActUnresolved, each with a different cost and potential outcome. For any planning state $S$, let $b(S)$ denote the current-label correctness probability obtained from its joint posterior $P$ using~\eqref{eq:joint-marginals}. When $S=S_t$, we have $b(S_t)=b_t$. Unlike \ActAccept, which retains the current label, \ActArbitrate obtains a new label whose correctness may also be affected by shared bias and the associated bias cues. Under the observation model in~\eqref{eq:unified-obs}, arbitration produces a correct label when its observation agrees with the held label $e=+1$ if $z=1$, or disagrees with the held label $e=-1$ if $z=0$. By weighting these two possibilities according to the current joint posterior $P$, we obtain the probability that \ActArbitrate produces a correct label at state $S$:
\begin{equation}
\small
\begin{aligned}
\alpha_{\ActD}(S;\mathbf x)
&=\sum_{\xi\in\{0,1\}^K}
P(1,\xi)\,
\sigma\bigl(w_{\ActD}+\Delta_t(\ActD,\xi)\bigr)\\
&\quad+\sum_{\xi\in\{0,1\}^K}
P(0,\xi)\,
\sigma\bigl(w_{\ActD}-\Delta_t(\ActD,\xi)\bigr).
\end{aligned}
\label{eq:alpha-D}
\end{equation}
For each state $S$, $\Delta_t(\ActD,\xi)$ is evaluated using the held label in $S$ and the task attributes $\mathbf x$. If \ActRelabel changes the held label, the cue directions are recomputed relative to the new label in the successor state. Since model parameters and $\mathbf x$ remain fixed within the task, we abbreviate $\alpha_{\ActD}(S;\mathbf x)$ as $\alpha_{\ActD}(S)$ when the context is clear. Subsequently, we define the expected losses of the three terminal actions as
\begin{equation}
\hspace{-3.5mm}\resizebox{0.92\columnwidth}{!}{$
\begin{aligned}
L_{\ActA}(S)&=c_{\mathrm{err}}\bigl(1-b(S)\bigr),~~~~~
L_{\ActD}(S)=c_{\ActD}+c_{\mathrm{err}}\bigl(1-\alpha_{\ActD}(S)\bigr),\\
L_{\ActU}(S)&=c_{\ActU},~~~~~
J_{\mathrm{term}}(S)=\min\{L_{\ActA}(S),L_{\ActD}(S),L_{\ActU}(S)\}.
\end{aligned}
$}\hspace{-3.5mm}
\label{eq:term-loss}
\end{equation}
Here, $L_{\ActA}(S)$ captures the expected error cost of accepting the current label, $L_{\ActD}(S)$ combines the arbitration cost with the expected error cost of its returned label, and $L_{\ActU}(S)$ represents the penalty for leaving the task unresolved. Accordingly, $J_{\mathrm{term}}(S)$ gives the minimum expected loss achievable by terminating QC. This quantity serves as the baseline against which continuing actions are evaluated in Step~2.

\smallskip\noindent\textbullet~\textbf{Step 2: Recursive evaluation of the expected remaining loss.}
Unlike terminal actions, the expected loss of \ActInspect or \ActRelabel depends on both the newly acquired observation and the decisions that can be made afterward. Therefore, evaluating these continuing actions requires accounting for their possible outcomes and the optimal decisions available at the corresponding successor states.

To capture these future decisions, we consider a finite planning horizon extended from the current state to the maximum number of continuing QC steps $T_{\max}$. For a planning state $S=(P,\hat y,C,t)$, we define the remaining planning depth as $h(S)=T_{\max}-t$, which specifies the maximum number of continuing actions that can still be executed.
Let $V_h(S)$ denote the minimum expected loss achievable from state $S$ within a remaining depth $h$, where $0\le h\le h(S)$. At depth $0$, no continuing action is allowed, so the immediate-termination loss is used. For $h\ge1$, the minimum expected remaining loss is obtained by evaluating all feasible actions. Accordingly, we obtain the Bellman recursion as
\begin{equation}
\begin{aligned}
V_0(S)&=J_{\mathrm{term}}(S),\\
V_h(S)
&=\min_{a\in\mathcal A_t^{\mathrm{feas}}}
Q_h(S,a),\qquad h\ge1,
\end{aligned}
\label{eq:bellman-V}
\end{equation}
where $t$ is the fourth component of $S$, and $\mathcal A_t^{\mathrm{feas}}$ is given by~\eqref{eq:A-feas}. Here, $Q_h(S,a)$ denotes the expected remaining loss when action $a$ is selected first at state $S$, followed by optimal decisions at subsequent steps. Specifically,
\begin{equation}
\hspace{-3.5mm}\resizebox{0.92\columnwidth}{!}{$
\begin{aligned}
Q_h(S,a)
&=L_a(S), &&a\in\{\ActA,\ActD,\ActU\},\\
Q_h(S,a)
&=c_a{+} \hspace{-3mm}\sum_{e\in\{+1,-1\}}
\hspace{-4mm}\Pr(e\mid S,a;\mathbf x)
V_{h-1}\!\left(\tau(S,a,e)\right),
&&a\in\{\ActI,\ActR\}.
\end{aligned}
$}\hspace{-3.5mm}
\label{eq:Q-cont}
\end{equation}
For a terminal action $a$, $Q_h(S,a)$ captures its immediate expected loss $L_a(S)$, since no further decisions are made. In contrast, for a continuing action $a$, the planner adds its execution cost $c_a$ to the expected loss of subsequent decisions. In~\eqref{eq:Q-cont}, this expectation is computed by weighting the minimum achievable loss $V_{h-1}$ at each successor state by the corresponding observation probability obtained in Phase~1. In this way, the recursion accounts for both the immediate cost of acquiring evidence and its potential to improve future QC decisions.

\smallskip\noindent\textbullet~\textbf{Step 3: Selecting the current action.}
Having evaluated the expected losses of all feasible actions, the planner selects the action that minimizes the expected remaining QC loss: at the current state $S_t$, the planner considers the full remaining planning depth $T_{\max}-t$ and determines the optimal action 
\begin{equation}
a_t^\star
=
\arg\!\min_{a\in\mathcal A_t^{\mathrm{feas}}}
Q_{T_{\max}-t}(S_t,a),
\label{eq:routing-rule}
\end{equation}
where $\mathcal A_t^{\mathrm{feas}}$ is given by~\eqref{eq:A-feas}. If multiple actions attain the same expected loss, we apply the tie-breaking order $\ActU\succ\ActA\succ\ActD\succ\ActI\succ\ActR$ to ensure deterministic action selection. This rule applies only when the expected losses are exactly equal and therefore does not affect the optimal expected loss. The selected action $a_t^\star$ is then executed at the current state $S_t$.

\subsection{Phase 3: Online Execution and Replanning}
\label{subsec:overall-procedure}

Phase~3 executes the action selected in Phase~2 and updates the online state based on its outcome. Specifically, if the selected action $a_t^\star$ is terminal, the QC episode ends and produces the corresponding output defined in Section~\ref{sec:problem-formulation}. Otherwise, the platform executes \ActInspect or \ActRelabel, observes the resulting outcome $e_t$, and updates the online state according to $S_{t+1}=\tau(S_t,a_t^\star,e_t)$. The planner then repeats the decision-making process from the updated state $S_{t+1}$. Importantly, only the successor state corresponding to the actual observation is retained, while the unrealized branches are discarded. This process continues until a terminal action is selected, as summarized in Algorithm~\ref{alg:jaber-routing}.

\begin{algorithm}[t]
\caption{JABER Finite-Horizon Routing}
\label{alg:jaber-routing}
\footnotesize
\DontPrintSemicolon
\SetAlgoNlRelativeSize{-1}
\SetKwInput{Input}{Input}
\SetKwProg{Fn}{Function}{:}{}
\SetKw{Ret}{return}
\Input{initial state $S_0=(P_0,\hat y_{\mathrm{init}},0,0)$, task context $\mathbf x$, and fixed model, loss, and budget parameters}
\textbf{Initialization:} $S\gets S_0$, $a^\star\gets\bot$\;
\While{$a^\star\notin\mathcal A_{\mathrm{term}}$}{
    $(\_,a^\star)\gets\textsc{Plan}(S,h(S),\mathbf x)$\;
    \If{$a^\star\in\{\ActI,\ActR\}$}{
        Execute $a^\star$ and observe $e$\;
        $S\gets\tau(S,a^\star,e)$ by \eqref{eq:state-transition}\;
    }
}
Read $(P,\hat y,C,t)$ from state $S$\;
\eIf{$a^\star=\ActA$}{
    $\hat y_{\mathrm{final}}\gets\hat y$\;
}{
    \eIf{$a^\star=\ActD$}{
        Execute \ActArbitrate and obtain $\hat y_{\mathrm{final}}$\;
    }{
        $\hat y_{\mathrm{final}}\gets\varnothing$\;
    }
}
$T_{\mathrm{stop}}\gets t$, $a_{T_{\mathrm{stop}}}\gets a^\star$, $S_{T_{\mathrm{stop}}}\gets S$\;
\textbf{Return:} $(\hat y_{\mathrm{final}},a_{T_{\mathrm{stop}}},S_{T_{\mathrm{stop}}})$\;
\BlankLine
\Fn{\textnormal{\textsc{Plan}}$(S,h,\mathbf x)$}{
    Read $(P,\hat y,C,t)$ from state $S$\;
    $\mathcal F\gets\mathcal A_{\mathrm{term}}$ if $h=0$; otherwise $\mathcal F\gets\mathcal A_t^{\mathrm{feas}}$ by \eqref{eq:A-feas}\;
    \For{$a\in\mathcal F\cap\mathcal A_{\mathrm{term}}$}{
        $Q[a]\gets L_a(S)$\;
    }
    \For{$a\in\mathcal F\cap\{\ActI,\ActR\}$}{
        $Q[a]\gets c_a$\;
        \For{$e\in\{+1,-1\}$}{
            $p_e\gets\Pr(e\mid S,a;\mathbf x)$ by \eqref{eq:predictive-e}\;
            $S'\gets\tau(S,a,e)$ by \eqref{eq:state-transition}\;
            $(v_e,\_)\gets\textsc{Plan}(S',h-1,\mathbf x)$\;
            $Q[a]\gets Q[a]+p_e v_e$\;
        }
    }
    $a^\star\gets\arg\min_{a\in\mathcal F}Q[a]$ with tie-breaking order $\ActU\succ\ActA\succ\ActD\succ\ActI\succ\ActR$\;
    \Ret $(Q[a^\star],a^\star)$\;
}

\end{algorithm}

The proposed procedure is guaranteed to terminate after a finite number of continuing actions. Specifically, each \ActInspect or \ActRelabel action incurs a cost of at least $c_{\min}=\min\{c_{\ActI},c_{\ActR}\}>0$, while no continuing action is feasible when $t\ge T_{\max}$. Therefore, the total number of executed \ActInspect and \ActRelabel actions is bounded by
$
\min\!\left\{T_{\max},\left\lfloor \frac{C_{\max}}{c_{\min}} \right\rfloor\right\}.
$

\subsection{Theoretical Analysis}
\label{subsec:theoretical-properties}

In this section, we theoretically investigate the limitations of repeated \ActInspect actions in
distinguishing label correctness from shared bias, establish conditions under which \ActRelabel
provides additional information to resolve this ambiguity, and show why the marginal beliefs
of correctness and shared bias alone are insufficient for optimal action selection. To this end, we
first focus on a single bias type to characterize the distinguishability of competing explanations
and then examine how the full joint posterior influences QC decisions. All the proofs are provided in Appendix~\ref{app:proofs}.

\smallskip\noindent\textbf{Notation and setup.}
Repeated observations supporting the held label can have two competing explanations: either the
label is correct, or it is incorrect but repeatedly supported due to persistent shared bias. To
formalize this ambiguity, we say that two latent states are observationally equivalent under action
$a$ and belong to the same alias class if they produce the same observation distribution for a fixed
task context $(\mathbf x,\hat y_t)$. We denote this equivalence by $(z,\xi)\sim_a(z',\xi')$.

To examine this ambiguity, we focus on a single bias type $k$ with fixed cue strength $r\neq0$,
setting $r_{t,k}=r$ and $r_{t,j}=0$ for all $j\neq k$. Letting $u=\xi_k$, we marginalize over the
remaining bias dimensions to obtain
\begin{equation}
P_{t,k}(z,u)
:= \hspace{-2mm}
\sum_{\substack{\xi\in\{0,1\}^K,~~ \xi_k=u}}P_t(z,\xi),
\qquad z,u\in\{0,1\}.
\label{eq:single-regime-marginal}
\end{equation}
Here, the first coordinate $z$ indicates whether the held label is correct, while the second
coordinate $u$ indicates whether bias type $k$ is active: arguments $(0,0)$ and $(0,1)$
represent an incorrect label with inactive and active bias, respectively, whereas $(1,0)$ and
$(1,1)$ represent a correct label under the respective bias states. The associated marginal
beliefs are
\[
\begin{aligned}
b_t&=P_{t,k}(1,0)+P_{t,k}(1,1),\\
\rho_{t,k}&=P_{t,k}(0,1)+P_{t,k}(1,1).
\end{aligned}
\]
The following results presume Assumptions~\ref{asmp:static-xi}
and~\ref{asmp:cond-indep-z-xi} with known model parameters. The repeated-\ActInspect analysis
additionally assumes the factorized initial prior in~\eqref{eq:joint-prior}.

For the selected bias type $k$, we define $d_a=\varphi_{a,k}r$ as the bias-induced shift in the
observation logit for action $a\in\{\ActI,\ActR,\ActD\}$. Under this setting, the logit
in~\eqref{eq:unified-obs} simplifies to $(2z-1)w_a+ud_a$. We focus on two competing explanations:
$(0,1)$, representing an incorrect label supported by active bias, and $(1,0)$, representing a
correct label with inactive bias. When $d_{\ActI}=2w_{\ActI}$, the logit under $(0,1)$ is
$-w_{\ActI}+2w_{\ActI}=w_{\ActI}$, so both explanations yield the same \ActInspect support
probability $\sigma(w_{\ActI})$, making them observationally equivalent under \ActInspect, i.e.,
$(0,1)\sim_{\ActI}(1,0)$. We refer to this condition as the \textit{exact aliasing point}. To examine the
consequences of ignoring shared bias, we compare our joint-belief approach against
conditional-independence (CI) updating, which disregards the shared-bias state and updates the
correctness belief using \ActInspect observations as if they were conditionally independent given
label correctness.

\begin{lemma}[Overconfidence under CI Updating]
\label{lem:ci-overconfidence}
If the true configuration is $(z,u)=(0,1)$, \ActInspect is repeated indefinitely, and
$d_{\ActI}>w_{\ActI}$, then (a.s. stands for \textit{almost surely})
\[
b_t^{\mathrm{CI}}\xrightarrow{\mathrm{a.s.}}1,
\]
where $b_t^{\mathrm{CI}}$ denotes the probability assigned by CI updating to the held label being
correct after $t$ observations.
\end{lemma}

The above result shows that when
the influence of shared bias exceeds the baseline correctness signal, CI updating can become
increasingly confident that an incorrect label is correct. Specifically, repeated \ActInspect
observations are interpreted as independent evidence of correctness, even when their agreement is
driven by persistent shared bias. Consequently, acquiring more judgments can reinforce an incorrect
belief rather than improve its accuracy.

\begin{proposition}[Repeated \ActInspect Preserves Posterior Odds within an Alias Class]
\label{prop:aliasing-noncollapse}
At $d_{\ActI}=2w_{\ActI}$, suppose the true configuration is $(0,1)$ and
$b_0,\rho_{0,k}\in(0,1)$. Repeated \ActInspect preserves the ratio of the two aliased posterior
probabilities for every $t$:
\begin{equation}
{P_{t,k}(1,0)}/{P_{t,k}(0,1)}
=
{P_{0,k}(1,0)}/{P_{0,k}(0,1)}.
\label{eq:within-class-ratio}
\end{equation}
Further, the posterior probabilities of the other two explanations converge almost surely to
zero. Thus, the correctness belief satisfies
\begin{equation}
b_t \xrightarrow{\mathrm{a.s.}} b_{\infty}^{\mathrm{alias}}
:=
\frac{b_0(1-\rho_{0,k})}
{b_0(1-\rho_{0,k})+(1-b_0)\rho_{0,k}}
\in(0,1).
\label{eq:prop3-pi}
\end{equation}
\end{proposition}
Unlike CI updating, Proposition~\ref{prop:aliasing-noncollapse} shows that the exact joint posterior
preserves uncertainty between explanations that \ActInspect cannot distinguish. In particular, at
the exact aliasing point, an incorrect label supported by active bias and a correct label without
active bias produce identical \ActInspect observation distributions. Therefore, regardless of how
many inspections are performed, their relative posterior probabilities remain unchanged, and the
correctness belief converges to a value strictly between zero and one rather than becoming falsely
certain. Moreover, near the aliasing point, the information gained per \ActInspect grows
quadratically with the deviation from exact aliasing, indicating that closely matched explanations
may remain difficult to distinguish within a QC horizon (see
Appendix~\ref{app:proof-aliasing-noncollapse}). Together, these results reveal two distinct
limitations of repeated inspection: ignoring shared bias can produce false certainty in an incorrect
label (Lemma~\ref{lem:ci-overconfidence}), whereas explicitly accounting for shared bias preserves
uncertainty when the available evidence cannot distinguish competing explanations
(Proposition~\ref{prop:aliasing-noncollapse}). This motivates considering alternative QC actions
that may provide more discriminative evidence.

To assess whether \ActRelabel can distinguish the two competing explanations that \ActInspect
cannot, we assume both have positive posterior probability and define the log posterior odds as 
$\Lambda_t=\log\!\bigl(P_{t,k}(0,1)/P_{t,k}(1,0)\bigr)$, whose increase indicates stronger evidence
in favor of the incorrect label supported by active bias. Let $\widetilde\Lambda_{t+1}$ denote the
log posterior odds after incorporating one \ActRelabel observation but before switching the held
label. Under the two competing explanations, the \ActRelabel observation distributions are
$p_{\ActR}^{\mathrm{bias}}=\mathrm{Bern}(\sigma(-w_{\ActR}+d_{\ActR}))$ and
$p_{\ActR}^{\mathrm{correct}}=\mathrm{Bern}(\sigma(w_{\ActR}))$. At the \ActInspect aliasing
point, these distributions are identical if and only if $d_{\ActR}=2w_{\ActR}$, equivalently
$w_{\ActI}\varphi_{\ActR,k}=w_{\ActR}\varphi_{\ActI,k}$. Thus, even when \ActInspect cannot distinguish the two explanations, \ActRelabel can provide discriminative evidence if its observation probabilities differ between an incorrect label supported by active bias and a correct label without active bias.

\begin{proposition}[\ActRelabel Provides Positive Expected Discriminative Evidence]
\label{prop:dealiasing-drift}
At the exact aliasing point, suppose that the true configuration is $(0,1)$ and
$w_{\ActI}\varphi_{\ActR,k}\neq w_{\ActR}\varphi_{\ActI,k}$. Then $(0,1)\not\sim_{\ActR}(1,0)$, and
the increment in the log posterior odds under the true observation distribution, before
label switching, satisfies
\begin{equation}
\hspace{-3.5mm}\resizebox{0.88\columnwidth}{!}{$
\mathbb E_{p_{\ActR}^{\mathrm{bias}}}\!\left[
\widetilde\Lambda_{t+1}-\Lambda_t
\,\middle|\,a_t=\ActR
\right]=D_{\mathrm{KL}}\!\left(
p_{\ActR}^{\mathrm{bias}}\middle\|p_{\ActR}^{\mathrm{correct}}
\right)>0.
$}\hspace{-3.5mm}
\label{eq:prop3-drift}
\end{equation}
\end{proposition}
Proposition~\ref{prop:dealiasing-drift} establishes that \ActRelabel can provide information that
repeated \ActInspect actions cannot obtain. The KL divergence measures the difference between the
two \ActRelabel observation distributions; its positive value means that the evidence favors the
true explanation on average. However, this additional information comes at an acquisition cost and
does not necessarily justify performing \ActRelabel. JABER thus evaluates whether its potential
benefits for subsequent QC decisions outweigh its cost through the expected total loss.

The optimal action also depends on the current joint posterior, particularly how the probabilities
are distributed across different combinations of label correctness and shared-bias states. This
raises an important question: \textit{is it sufficient to maintain separate probabilities of label
correctness and bias activity, or must their joint dependence also be preserved?} The following
proposition addresses this question.

\begin{proposition}[Insufficiency of Correctness and Bias Marginals for Optimal Action Selection]
\label{prop:marginal-insufficiency}
Under the same observation model, costs, budget, and remaining horizon, there exist two interior
joint beliefs (positive on every joint state) over $(z,\xi)$ with identical $b$ and $\rho$ but
different optimal first actions.
\end{proposition}

Proposition~\ref{prop:marginal-insufficiency} demonstrates that the optimal QC action depends not
only on the individual probabilities of label correctness and shared-bias activity but also on the
statistical dependence between these two variables. To illustrate this result,
Appendix~\ref{app:proof-marginal-insufficiency} constructs an example with a single bias type $K=1$
and considers two different joint posterior distributions $P(z,\xi)$ over label correctness and
shared-bias activity. Although these two beliefs assign the same marginal probabilities to label
correctness and bias activity, i.e., $b=\rho=0.5$, they differ in how their probability mass is
distributed across the joint states. In the first belief, most of the probability mass is
concentrated on $(0,1)$ and $(1,0)$, which \ActInspect cannot distinguish, making \ActRelabel the
preferred first action. In contrast, the second belief concentrates most of its probability mass on
$(0,0)$ and $(1,1)$, which \ActInspect can distinguish, making \ActInspect the preferred first
action. This distinction is lost under Marginal-Projection, which approximates the joint posterior
by the product of its marginals ($P(z,\xi)\mapsto P(z)P(\xi)$ for $K=1$). Consequently, both joint
beliefs are reduced to the same approximate distribution, preventing the planner from distinguishing
situations that require different QC actions.

Appendix~\ref{app:proof-planner-optimality} establishes the optimality of the resulting policy under
the specified observation model and resource constraints. It further shows that increasing planning
depth cannot increase the minimum expected loss, since the planner can either exploit these
additional opportunities to improve its decisions or terminate earlier when further evidence
acquisition is not worthwhile. Section~\ref{sec:experiments} empirically examines the benefits of
such multistep decision-making by comparing JABER against a one-step lookahead approach.

\section{Experiments}
\label{sec:experiments}

In the following, we first conduct controlled synthetic experiments using observations generated
from the mathematical model to evaluate joint-posterior representation and multistep planning of
\textit{JABER}. Through a controlled real-API workflow on \textit{FEVER}~\cite{thorne2018fever}, we then examine QC decisions under
persistent misleading information. Finally, through a set of supplementary experiments, we examine additional method comparisons
and model misspecification, and analyze recorded judgments from \textit{HealthContradict}~\cite{zhang2026healthcontradict} and four
additional public datasets to assess how judgment informativeness, reliability, and costs affect
policy choices.

\subsection{Comparative Methods and Evaluation Protocol}
\label{subsec:exp-setup}

We consider the following three ablation methods:
\vspace{-.75mm}

\smallskip\noindent\textbullet~\textit{Marginal-Projection:} 
This method evaluates the importance of preserving posterior dependence. It uses \textit{JABER}'s observation model, Bayesian updates, and finite-horizon planner but replaces the updated joint posterior with the product of its variable-wise marginals, preserving individual probabilities while discarding their dependencies.

\vspace{-.75mm}

\smallskip\noindent\textbullet~\textit{CI Updating:} 
This method evaluates the importance of modeling persistent shared bias. It omits the bias variables and treats successive judgments as conditionally independent given label correctness, preventing it from explicitly distinguishing genuine correctness from bias-induced agreement.

\vspace{-.75mm}

\smallskip\noindent\textbullet~\textit{One-step lookahead:} 
This method evaluates the benefit of multistep planning. It retains \textit{JABER}'s joint-belief updates but considers only one continuing action (\ActInspect or \ActRelabel), followed by the optimal terminal decision.


To conduct comparisons, 
we consider task-adapted \textit{CLARA-based}~\cite{nguyen2020clara} and \textit{Raykar-style}~\cite{raykar2014sequential} policies, which assume conditionally independent reviewer judgments. For fair comparison, these methods and \textit{JABER} operate under a matched \textit{Query/Stop} setting, allowing additional judgments or termination with a decision on the held label's correctness. To evaluate alternative dependence models, we include \textit{Full Ising}~\cite{ising2026aggregation}, which captures pairwise judgment correlations using \textit{JABER}'s sequential planner.



We further include fixed-query, \textit{Always-Accept}, and \textit{Always-Relabel} rules to compare against simpler QC strategies. Finally, \textit{Best Precommitted Sequence} evaluates the benefit of adaptive action selection: it uses JABER's joint-belief updates but fixes the order of \ActInspect and \ActRelabel before acquiring observations, while permitting early termination. Among all sequences satisfying the resource constraints, we select the one with the lowest model-expected QC loss.

\smallskip\noindent\textbf{Language Models.}
The controlled synthetic experiments generate observations from the mathematical model and do not
involve LLM calls. In the real-API \textit{FEVER} experiments, \textit{Qwen-Flash}
(\texttt{qwen-flash-}\allowbreak\texttt{2025-07-28}) provides repeated judgments for \ActInspect{},
\textit{Qwen-Plus} (\texttt{qwen-plus-}\allowbreak\texttt{2025-12-01}) provides replacement labels
for \ActRelabel{}, and \textit{DeepSeek} (\texttt{deepseek-}\allowbreak\texttt{v4.1-flash}) performs
independent evidence retrieval and final verification for \ActArbitrate{}. The
\textit{HealthContradict} experiments use the same \textit{Qwen-Flash} version for the initial and
subsequent judgments. The remaining public-dataset diagnostics analyze existing human annotations
and model-judgment records.

\smallskip\noindent\textbf{Metrics.}
Our primary metric is \textit{mean total QC loss} (the objective function of
$\bm{\mathcal P}_0$), which combines action costs, erroneous-output penalties, and unresolved-task penalties. We also report the \textit{false-accept rate} (episodes accepting an incorrect  label) and \textit{erroneous-output rate} (episodes producing any incorrect final label), both normalized by the total number of episodes. Synthetic experiments compute expected losses through full action–observation path enumeration, while real-judgment experiments report mean realized losses across samples. Confidence intervals use paired bootstrap resampling.

\subsection{Controlled Synthetic Experiments}
\label{subsec:exp-simulation}


We conduct controlled synthetic experiments to evaluate two core components of \textit{JABER}: (i) preserving the joint posterior over label correctness and shared bias, and (ii) using this posterior for finite-horizon multistep QC planning.

\smallskip\noindent\textbf{Setup.} We generate QC observations using the probabilistic model in~\eqref{eq:unified-obs}, with known model parameters, and consider initial correctness probabilities $b_0\in\{0.50,0.60,0.70\}$. We normalize costs by setting $c_I=1$ and consider two penalty settings to compute~\eqref{eq:episode-loss}: \textit{Original}, with $(c_{\mathrm{err}},c_\mathrm{U})=(20,10)$, and \textit{Risk-sensitive}, with $(c_{\mathrm{err}},c_\mathrm{U})=(80,20)$, which assigns greater penalties to incorrect final labels.

We first consider two shared-bias types ($K=2$), with different bias-coupling strengths for \ActInspect and \ActRelabel (i.e., the action-specific bias-coupling coefficients $\varphi_{a,k}$ defined in~\eqref{eq:Delta-def}, the values of which are reported in Appendix~D.1). Specifically, \ActInspect is more susceptible to the first bias, whereas \ActRelabel is more susceptible to the second, allowing us to examine the benefits of selecting actions with different sensitivities to shared bias. We also consider $K=3$, where an additional bias influences both actions, to evaluate the effect of multiple shared-bias sources.



\smallskip\noindent\textbf{Joint-belief decision gains.}
Table~\ref{tab:exp-core-joint} compares \textit{JABER} with \textit{Marginal-Projection} and \textit{CI Updating} under the aforementioned cost settings. Overall, \textit{JABER} achieves the lowest (or tied-lowest) expected QC loss in all configurations, demonstrating the benefits of explicitly modeling shared bias and preserving posterior dependence. Also, under the \textit{Risk-sensitive} setting, it consistently achieves lower false-accept rates than both baselines across all six combinations of $b_0$ and $K$. We note that, under the \textit{Original} setting, where erroneous outputs incur a lower penalty, \textit{JABER} prioritizes reducing QC costs, achieving lower expected loss than \textit{Marginal-Projection} in five configurations and matching it in one, albeit with higher false-accept rates in the five improved configurations. These results illustrate how joint-belief modeling enables cost-aware QC decisions that adapt to the relative penalties of annotation errors and resource consumption (i.e., the competing terms captured by~\eqref{eq:episode-loss}).

\begin{table}[t]
\centering
\caption{Expected total QC loss and false-accept rate for \textit{JABER} and its updating ablations
in controlled synthetic experiments. Bold entries mark the lowest displayed value for each metric.
\(K\) denotes the number of shared-bias types, and \(b_0\) is the initial correctness probability.}
\label{tab:exp-core-joint}
\vspace{-3mm}
\footnotesize
\setlength{\tabcolsep}{1.7pt}
\renewcommand{\arraystretch}{1.03}
\begin{tabular*}{\columnwidth}{@{\extracolsep{\fill}}ccrrrrrr@{}}
\toprule
$K$ & $b_0$ & \multicolumn{2}{c}{\textit{JABER}} & \multicolumn{2}{c}{\makecell{\textit{Marginal-}\\\textit{Projection}}} & \multicolumn{2}{c}{\makecell{\textit{CI}\\\textit{Updating}}}\\
\cmidrule(lr){3-4}\cmidrule(lr){5-6}\cmidrule(l){7-8}
& & Loss $\downarrow$ & \makecell{False accept\\(\%) $\downarrow$} & Loss $\downarrow$ & \makecell{False accept\\(\%) $\downarrow$} & Loss $\downarrow$ & \makecell{False accept\\(\%) $\downarrow$}\\
\midrule
\multicolumn{8}{@{}l}{\textit{Original}}\\
2 & 0.50 & \textbf{7.95} & 26.43 & 8.29 & \textbf{21.48} & 8.22 & 22.14\\
2 & 0.60 & \textbf{7.24} & 23.52 & 7.30 & \textbf{22.43} & \textbf{7.24} & 23.52\\
2 & 0.70 & \textbf{6.00} & 30.00 & 6.25 & \textbf{17.82} & 6.31 & 18.51\\
\addlinespace[2pt]
3 & 0.50 & \textbf{8.23} & 28.51 & 8.57 & \textbf{23.59} & 8.54 & 24.37\\
3 & 0.60 & \textbf{7.37} & 24.82 & 7.49 & \textbf{24.28} & \textbf{7.37} & 24.82\\
3 & 0.70 & \textbf{6.00} & 30.00 & \textbf{6.00} & 30.00 & 6.36 & \textbf{19.60}\\
\midrule
\multicolumn{8}{@{}l}{\textit{Risk-sensitive}}\\
2 & 0.50 & \textbf{18.37} & \textbf{4.49} & 19.86 & 12.86 & 19.91 & 15.21\\
2 & 0.60 & \textbf{17.62} & \textbf{9.62} & 18.22 & 11.09 & 18.74 & 13.43\\
2 & 0.70 & \textbf{15.44} & \textbf{8.00} & 15.88 & 8.52 & 15.87 & 10.89\\
\addlinespace[2pt]
3 & 0.50 & \textbf{19.05} & \textbf{6.21} & 20.27 & 11.71 & 21.41 & 17.37\\
3 & 0.60 & \textbf{18.19} & \textbf{6.73} & 19.87 & 13.12 & 19.65 & 15.00\\
3 & 0.70 & \textbf{16.08} & \textbf{9.52} & 16.75 & 10.08 & 16.50 & 12.29\\
\bottomrule
\end{tabular*}
\vspace{-3mm}
\end{table}

\begin{figure}[t]
\centering
\vspace{-1mm}
\includegraphics[width=\columnwidth, trim= 0 0 0 9, clip]{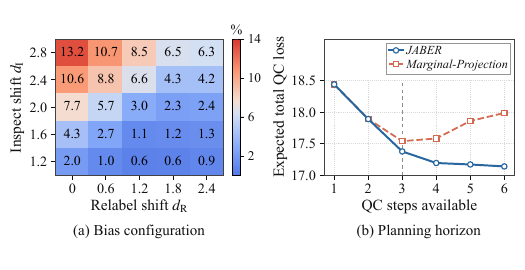}
\vspace{-13mm}
\caption{\textit{JABER}'s percentage reduction in expected total QC loss relative to
\textit{Marginal-Projection}; higher is better. Both panels use \textit{Risk-sensitive} costs and
acquisition budget 8. Each cell or point averages the loss reduction over
\(b_0\in\{0.50,0.60,0.70\}\). (a) Single-bias configurations vary the bias shifts affecting
\ActInspect{} and \ActRelabel{}, with at most six continuing steps. (b) A two-bias configuration
varies the available continuing steps; the dashed line marks the first nonzero gain at three steps.}
\label{fig:exp-controlled-summary}
\vspace{-5mm}
\end{figure}

We further examine how action-specific bias effects influence these gains. Fig.~\ref{fig:exp-controlled-summary}(a) compares \textit{JABER} and \textit{Marginal-Projection} across 25 configurations of bias-induced shifts (i.e., varying values of $d_a$, $a\in\{\ActI,\ActR\}$,  defined in Section~\ref{subsec:theoretical-properties}) for $K=1$. 
Under the \textit{Risk-sensitive} cost setting, we compute the relative reduction in expected QC loss for each configuration, averaging equally across the three initial correctness probabilities $b_0\in\{0.50,0.60,0.70\}$.
\textit{JABER} reduces expected QC loss by an average of 4.65\% (median 4.24\%), with improvements ranging from 0.6\% to 13.2\% across all configurations. The gains are generally larger when shared bias makes \ActInspect{} less informative while \ActRelabel{} remains comparatively discriminative, consistent with the aliasing and distinguishability analysis in Section~\ref{subsec:theoretical-properties}.

\smallskip\noindent\textbf{Effect of planning horizon on joint-belief gains.}
Fig.~\ref{fig:exp-controlled-summary}(b) compares \textit{JABER} and \textit{Marginal-Projection} for $K=2$ under the same observation model and evidence-acquisition budget, while varying the maximum number of continuing actions $T_{\max}$. Both methods use the full available planning horizon. Their expected losses coincide for $T_{\max}\leq 2$, but \textit{JABER} achieves progressively larger gains thereafter, with its average loss reduction increasing from 0.93\% at $T_{\max}=3$ to 4.51\% at $T_{\max}=6$. These results indicate that preserving posterior dependence becomes increasingly valuable when more evidence-acquisition opportunities are available.

\subsection{Controlled Real-API QC on \textit{FEVER}}
\label{subsec:exp-fever}


\textit{FEVER} is a fact-verification dataset containing claims labeled as \textit{Supported}, \textit{Refuted}, or \textit{NotEnoughInfo} based on Wikipedia evidence~\cite{thorne2018fever}. We use its \textit{Supported}/\textit{Refuted} subset to evaluate \textit{JABER} in a real-application programming interface (real-API) QC workflow, where multiple LLM judgments may be influenced by the same misleading background information, leading to correlated annotation errors.

\smallskip\noindent\textbf{Setup.} We calibrate the observation model using 16 development claims (i.e., claims used to estimate the model parameters) and select the primary experimental configuration by comparing QC policies on these claims. After fixing the parameters, we test the methods on 40 separate claims with no overlap in claims or annotated Wikipedia evidence pages. To evaluate QC under both correct and incorrect initial annotations, we assign 32 correct and eight incorrect initial labels. Further, to induce correlated annotation errors, we construct a shared background account for each claim (i.e., a common description containing a factual premise and a provisional label based on that premise). Both \ActInspect{} and \ActRelabel{} receive this account, allowing their judgments to be influenced by the same potentially misleading information. Each QC episode follows one of two fixed review instructions: either independently verify the claim or rely on the background account unless explicit contradictory evidence is found. These instructions create different levels of susceptibility to the shared information (i.e., different tendencies to follow the background account even when it is misleading). The planner does not observe which review instruction is used. Instead, it uses the observed judgments to infer additional susceptibility to the shared background information, represented by a latent component of the shared-bias state $\boldsymbol{\xi}$ defined in Section~\ref{subsec:hidden-state}.
For QC execution, \ActInspect{} uses \textit{Qwen-Flash} to obtain repeated judgments with the same prompt at temperature 0.7, allowing variation across responses. \ActRelabel{} uses \textit{Qwen-Plus} to generate a replacement label based on the background account and retrieved evidence. In contrast, \ActArbitrate{} uses \textit{DeepSeek} to independently retrieve evidence and verify the claim without receiving the shared account, providing an alternative source of verification.

\textit{JABER} selects QC actions online (i.e., each action is chosen before observing its API response), while the comparison policies are evaluated by replaying previously recorded API judgments (this ensures that different policies receive the same judgments for identical actions, enabling a consistent comparison while reducing variability from repeated LLM calls). The evidence-acquisition budget limits the estimated costs of \ActInspect{} and \ActRelabel{}, while the cost of \ActArbitrate{} is accounted for separately because it terminates QC. The reported QC loss includes actual API fees incurred during execution, even when they exceed the estimated budget. We also consider two simpler baselines: \textit{Cheap Flash Relabel}, which directly outputs a \textit{Qwen-Flash} judgment based only on retrieved evidence without further QC, and the \textit{development-selected fixed escalation rule}, which first performs \ActRelabel{} and then invokes \ActArbitrate{} if the replacement label disagrees with the initial label.

\begin{figure}[t]
\centering
\vspace{-1mm}
\includegraphics[width=\columnwidth, trim = 0 0 0 3, clip]{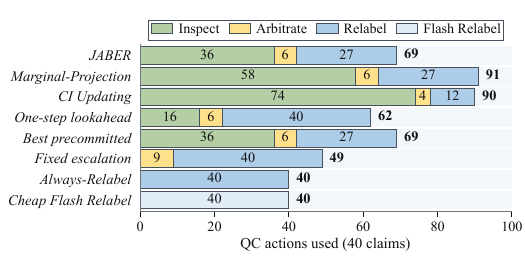}
\vspace{-10.75mm}
\caption{QC action counts on the 40 primary \textit{FEVER} evaluation claims. Segment labels give
counts by action; labels at bar ends give totals. \ActAccept{} requires no API call and is omitted.
\textit{Fixed escalation} denotes the development-selected rule in Table~\ref{tab:fever-primary}.}
\label{fig:fever-actions}
\vspace{-5mm}
\end{figure}

\smallskip\noindent\textbf{Results.} Table~\ref{tab:fever-primary} reports the results on 40 \textit{FEVER} claims. \textit{JABER}, \textit{Marginal-Projection}, \textit{One-step lookahead}, and the \textit{development-selected fixed escalation rule} all produce correct final labels without unresolved outcomes. However, \textit{JABER} achieves the lowest mean total QC loss of 0.00088455  Chinese yuan (CNY), reducing loss by 2.67\%, 16.99\%, and 30.45\% relative to these three baselines, respectively.
To examine these savings, Fig.~\ref{fig:fever-actions} compares the numbers of QC actions: \textit{JABER} performs 36 \ActInspect{} actions across the 40 claims, compared with 58 for \textit{Marginal-Projection}, while both methods perform the same numbers of \ActRelabel{} and \ActArbitrate{} actions and produce identical final labels. This indicates that preserving posterior dependence can help avoid redundant inspections without reducing labeling accuracy. Furthermore, while \textit{One-step lookahead} initially selects \ActRelabel{} for every claim, \textit{JABER} accepts the held label after one \ActInspect{} for 13 claims, illustrating how multistep planning can reduce evidence-acquisition costs. Fig.~\ref{fig:fever-actions} further shows that direct replacement baselines perform fewer QC actions, explaining their lower API fees in Table~\ref{tab:fever-primary}. However, this reduced verification comes at the expense of labeling accuracy, resulting in higher total QC losses under the primary cost setting, as also reported in Table~\ref{tab:fever-primary}.

To further investigate the sources of JABER's performance improvements, we compare it against \textit{Best Precommitted Sequence} to isolate the benefit of adaptively selecting the evidence-acquisition order.  This baseline uses JABER's joint posterior and finite-horizon planning but fixes the order of \ActInspect{} and \ActRelabel{} beforehand, while allowing early termination. As shown in Fig.~\ref{fig:fever-actions} and Table~\ref{tab:fever-primary}, both methods achieve identical action sequences, final labels, and total loss across all 40 claims. This finding implies that, in this setting, the reductions in QC loss relative to \textit{Marginal-Projection} and \textit{One-step Lookahead} arise from joint-belief modeling and multistep decision-making rather than dynamically adapting the evidence-acquisition order.

\begin{table}[t]
\centering
\caption{Primary \textit{FEVER} results on 40 evaluation claims. Errors count incorrect final
labels; no listed method leaves a claim unresolved. Mean API fees and total QC losses are in
\(10^{-3}\) Chinese yuan (CNY). Total loss adds a penalty of 0.058678 CNY per erroneous output to
API fees.}
\label{tab:fever-primary}
\vspace{-3mm}
\scriptsize
\setlength{\tabcolsep}{3pt}
\renewcommand{\arraystretch}{1.06}
\begin{tabularx}{\columnwidth}{@{}Xrrr@{}}
\toprule
 Method & \makecell{Errors $\downarrow$\\ / 40} & \makecell{Mean\\ API fee $\downarrow$} & \makecell{Mean total\\ QC loss $\downarrow$}\\
\midrule
\textit{JABER} & 0 & 0.8845 & 0.8845\\
\textit{Marginal-Projection} & 0 & 0.9088 & 0.9088\\
\textit{CI Updating} & 1 & 0.5837 & 2.0507\\
\textit{One-step lookahead} & 0 & 1.0656 & 1.0656\\
\textit{Best precommitted sequence} & 0 & 0.8845 & 0.8845\\
\textit{Development-selected fixed escalation rule} & 0 & 1.2719 & 1.2719\\
\textit{Always-Relabel} & 1 & 0.6738 & 2.1408\\
\textit{Cheap Flash Relabel} & 3 & 0.1090 & 4.5098\\
\bottomrule
\end{tabularx}
\vspace{-3mm}
\end{table}

\smallskip\noindent\textbf{Decision trajectory.}
To illustrate how \textit{JABER} makes sequential QC decisions
in practice, we examine a recorded decision trajectory
(i.e., the sequence of QC actions, observed judgments, and
belief updates) involving the publication date of
\emph{The Hobbit}, a fantasy novel by J. Tolkien.
The true publication date is September 21, 1937, and the evaluated claim incorrectly states that the book was not published on this date. Thus, the initial label \textit{Refuted} is correct. However, the shared background account incorrectly reports September 22, 1937, causing both \ActInspect{} and \ActRelabel{} to support the incorrect label \textit{Supported}. After the first \ActInspect{}, \textit{JABER} increases its posterior belief that the judgments are influenced by the shared background information. Subsequently, \ActRelabel{} changes the held label to \textit{Supported}, whose estimated correctness probability becomes 0.756 despite being incorrect. Rather than accepting this label, \textit{JABER} selects \ActArbitrate{}, which independently retrieves evidence and restores the correct label \textit{Refuted}. This trajectory illustrates how accounting for shared bias can motivate further verification even when the current label appears likely to be correct.

\subsection{Supplementary Experiments}
\label{subsec:exp-supplementary}
This subsection summarizes the supplementary experiments detailed in the appendix. For the synthetic experiments in Section~\ref{subsec:exp-simulation}, we provide additional comparisons of planning depth and sequential querying methods and evaluate robustness to observation-model misspecification. For the real-API \textit{FEVER} experiments in Section~\ref{subsec:exp-fever}, we investigate the sensitivity of QC performance to terminal penalties. Finally, experiments on \textit{HealthContradict} and other public datasets assess the informativeness of follow-up judgments, the reliability of replacement labels, and how these factors, together with action costs, influence \textit{JABER}'s performance relative to other QC strategies.


\smallskip\noindent\textbf{Planning depth.} We investigate the benefit of multistep planning while keeping
the maximum number of available QC steps fixed. We compare
\textit{JABER} against \textit{One-step Lookahead}, which uses
the same joint posterior and QC constraints but considers only
one \ActInspect{} or \ActRelabel{} action followed by the
optimal terminal decision, replanning after each observation.
The experimental setup and false-accept-rate results are
provided in Appendix~\ref{app:exp-query-stop}. In this separate
single-bias experiment, \textit{JABER} reduces expected QC
loss by 3.40\% and 10.05\% under the \textit{Original} and
\textit{Risk-sensitive} cost settings, respectively, relative to
\textit{One-step Lookahead}. These results demonstrate that accounting for multiple future QC decisions can reduce expected total loss compared with considering only the next evidence-acquisition action.

\smallskip\noindent\textbf{Comparison with sequential-query methods.}
To compare \textit{JABER} against sequential
querying methods, we use the matched \textit{Query/Stop}
setting introduced in Section~\ref{subsec:exp-setup},
where all methods operate under the same action set and
resource constraints. As shown in
Table~\ref{tab:exp-reference-baselines} of
Appendix~\ref{app:exp-query-stop}, \textit{JABER} achieves
lower expected total QC loss than the 
\textit{CLARA-based} and \textit{Raykar-style} policies
and fixed-query rules. However, fixed-query policies
requesting three or five judgments achieve lower
false-accept rates under the \textit{Original} cost setting.
These results highlight the advantage of adaptive evidence acquisition: while collecting more judgments can reduce false accepts, \textit{JABER} achieves lower expected total QC loss by accounting for both the reliability and cost of additional evidence when deciding whether to continue QC.


\smallskip\noindent\textbf{Observation-function misspecification.}
We evaluate the robustness of joint-belief modeling when the
true observation-generating process differs from the model
assumed by the planner. Specifically, we generate observations using \textit{probit}
(i.e., a probability function based on the standard normal
cumulative distribution) and centered complementary \textit{log-log}
functions (i.e., an asymmetric probability function based
on an exponential transformation), while both
\textit{JABER} and \textit{Marginal-Projection} retain
the sigmoid-based observation model in~\eqref{eq:unified-obs}.
As shown in Table~\ref{tab:exp-link-misspec} of
Appendix~\ref{app:exp-robustness}, \textit{JABER} reduces
expected QC loss by 2.30\%--2.63\% under the
\textit{Original} and 5.81\%--5.90\% under
\textit{Risk-sensitive} cost setting. These findings suggest that preserving posterior dependence can improve QC decisions even when the assumed observation model does not perfectly match the true judgment-generating process, highlighting JABER's robustness to the examined forms of model misspecification.


\smallskip\noindent\textbf{Terminal-loss sensitivity on \textit{FEVER}.}
To further characterize the operating conditions under which
\textit{JABER} is most beneficial, we examine its sensitivity
to penalties for incorrect final labels and unresolved tasks.
Table~\ref{tab:fever-all-costs} in
Appendix~\ref{app:exp-fever-details} reports mean total QC
loss (in $10^{-3}$ CNY) under different terminal penalties,
with E/U denoting the numbers of erroneous final outputs
and unresolved tasks among the 40 claims, respectively.
The results reveal that \textit{JABER}'s relative performance
depends on the terminal-loss weights. Under reduced terminal
penalties, \textit{JABER} incurs 19.33\% higher mean QC loss
than \textit{Marginal-Projection}, despite having the same
number of erroneous outputs and fewer unresolved tasks.
When the erroneous-output penalty is doubled,
\textit{JABER} incurs 6.43\% higher loss than
\textit{Best Precommitted Sequence}, although both produce
correct final labels for all 40 claims. These findings reveal a limitation of joint-belief planning: its advantages under the primary cost setting do not necessarily persist when terminal penalties change. This is expected because changing these penalties alters the tradeoff between acquiring additional evidence and terminating QC. Consequently, a policy that achieves lower loss under one cost setting may become less cost-effective under another, highlighting the importance of evaluating QC methods under application-specific error penalties and resource costs.

\smallskip\noindent\textbf{Replacement cost on \textit{HealthContradict}.}
We use the \textit{HealthContradict} health-question dataset~\cite{zhang2026healthcontradict}
to investigate how the cost and reliability of replacement
judgments influence QC decisions. In the reference-assisted
setting, \ActRelabel{} receives a reference passage selected
using the gold label (i.e., the known correct answer) and
directly outputs a replacement label without further QC.
This controlled setting allows us to examine when obtaining
a reliable replacement is preferable to collecting additional
judgments. Fig.~\ref{fig:health-cost-appendix} in
Appendix~\ref{app:exp-health-details} shows how mean QC
loss changes with the \ActRelabel{} cost under the
\textit{Risk-sensitive} setting, with \ActInspect{} cost
normalized to 1. The reference-assisted replacement achieves
92.5\% accuracy, and near the measured API fee ratio of
1.013, all three methods (\textit{JABER},
\textit{Marginal-Projection}, and \textit{CI Updating})
select immediate \ActRelabel{}. Indeed, \textit{JABER}
matches \textit{Always-Relabel} for $c_{\ActR}=1$--$9$,
indicating that additional evidence acquisition provides
no reduction in QC loss when reliable replacement is
sufficiently inexpensive. At $c_{\ActR}=10$,
\textit{JABER} instead selects \ActUnresolved{} because
its calibrated expected replacement loss exceeds the
unresolved-task penalty. Further, across 60 combinations of erroneous-output penalties, unresolved-outcome penalties, and \ActRelabel{} costs (detailed in Appendix~\ref{app:exp-health-details}), \textit{JABER} achieves lower QC loss than \textit{Marginal-Projection} in 18 configurations and \textit{CI Updating} in 13, while matching their performance in 33 and 43 configurations, respectively. In six configurations, it outperforms both methods without increasing erroneous final outputs. 
These findings reveal an important operating condition:
when a reliable replacement label is available at low
cost, immediate replacement can be more cost-effective
than sequential inspection, limiting the additional
benefits of joint-belief planning.

\smallskip\noindent\textbf{Reference-selection dependence.}
To assess the dependence of \ActRelabel{} on reference
selection, we provide it with both conflicting passages
instead of a passage selected using the gold label.
This removes the advantage of having a reference
known to support the correct answer and makes
replacement judgments more challenging. Under this
setting, \ActRelabel{} achieves a smoothed calibration
accuracy of 72.7\%, causing its estimated expected
loss to exceed the \ActUnresolved{} penalty.
Consequently, \textit{JABER} selects \ActUnresolved{}
throughout the evaluation. However, the observed
\ActRelabel{} accuracy is 80\%, allowing
\textit{Always-Relabel} to achieve lower realized
QC loss at $c_{\ActR}=1$--$3$, as detailed in
Appendix~\ref{app:exp-health-details}.
These findings highlight a limitation of belief-based
QC planning: when action reliability is underestimated
during calibration, the planner may unnecessarily
abstain even when acquiring a replacement label would
be more cost-effective.

\smallskip\noindent\textbf{Repeated-judgment information on \textit{LLMBar}.}
We use \textit{LLMBar}, which evaluates LLM judgments against
human preferences~\cite{zeng2024llmbar}, to investigate whether
additional judgments provide useful evidence about the
correctness of initial labels. In the adversarial subset,
subsequent judgments support incorrect initial choices in
68.5\% of cases and correct initial choices in 68.8\%, as
detailed in Appendix~\ref{app:exp-external-details}.
These nearly identical support rates indicate that additional
agreement provides little information for distinguishing
correct from incorrect labels. Consequently, in the examined
replay setting, acquiring further \ActInspect{} judgments
incurs higher total QC loss than immediately accepting the
initial label. These findings reinforce a central motivation
of \textit{JABER}: repeated agreement should not be treated
as reliable evidence of correctness when the underlying
judgments provide little discriminative information.


\smallskip\noindent\textbf{Error detection and replacement on \textit{LLMJudge}.}
We use \textit{LLMJudge}, which provides human relevance
ratings as reference labels~\cite{rahmani2025judging},
to examine whether judgments that help detect incorrect
initial labels can also provide reliable replacements.
As detailed in Appendix~\ref{app:exp-external-details},
the mean follow-up support rate is 17.19 percentage points
higher for correct initial labels than for incorrect ones,
indicating that these judgments contain useful information
about label correctness. However, the replacement judgments
have a higher overall error rate than the initial labels.
Even with zero \ActRelabel{} cost, selectively switching labels
based on the observed judgment history reduces the error
rate by only 0.723 percentage points compared with a fixed
majority-vote rule (i.e., selecting the most frequent label
among the initial and three follow-up judgments), with a
95\% confidence interval of [-1.094, 2.437] percentage
points that includes zero. These findings reveal an important
distinction: although additional judgments may help
identify incorrect labels, they do not necessarily provide
reliable replacements. Thus, the value of \ActRelabel{}
depends not only on detecting errors but also on the
reliability of the newly acquired label.


\section{Conclusion and Future Work}

We presented JABER, a bias-aware framework for sequential quality control (QC) in human--AI
collaborative annotation. By jointly modeling label correctness and shared bias, JABER accounts for
dependencies among judgments and uses finite-horizon planning to select QC actions that balance
evidence acquisition costs, terminal error risks, and available resources. Our theoretical analysis
demonstrated how repeated agreement can lead to overconfidence in incorrect labels, established when
additional inspections fail to resolve uncertainty, and highlighted the importance of preserving the
joint posterior for effective action selection. Experiments showed that JABER
reduces expected QC loss versus Marginal-Projection and conditional-independence baselines,
especially when errors incur high costs. In the real-API FEVER workflow, JABER achieved
lower mean loss than Marginal-Projection and One-step lookahead in the primary setting, matching the
paths and loss of the best precommitted sequence with the same joint posterior. Analyses of
HealthContradict and four additional public datasets showed how evidence dependence,  reliability
of QC judgments, and costs can limit the benefits of further QC or favor simpler policies. Promising
directions for future research include extending the binary-label formulation to graded or
multidimensional feedback, dynamically learning annotator reliability and shared biases from incoming
judgments, and developing scalable belief-update and planning algorithms for large-scale annotation
systems. 

{\linespread{0.92}\tiny
\bibliographystyle{IEEEtran}
\bibliography{references}}

\clearpage
\appendices

\phantomsection
\pdfbookmark[1]{Appendix Contents}{appendix-contents}
\begin{center}
\textbf{Appendix Contents}
\end{center}
\begingroup
\setlength{\tabcolsep}{0pt}
\renewcommand{\arraystretch}{1.10}
\newcommand{\AppendixContentsSection}[2]{%
  \hyperref[#1]{\textbf{\ref*{#1}.\enspace #2}} &
  \hyperref[#1]{\pageref*{#1}}\\[0.35em]}
\newcommand{\AppendixContentsSubsection}[2]{%
  \hspace*{0.8em}\hyperref[#1]{\ref*{#1}\enspace #2} &
  \hyperref[#1]{\pageref*{#1}}\\[0.15em]}
\noindent\begin{tabularx}{\columnwidth}{@{}>{\raggedright\arraybackslash}X@{\hspace{1em}}r@{}}
\AppendixContentsSection{app:proofs}{Proofs of Theoretical Properties}
\AppendixContentsSubsection{app:proof-ci-overconfidence}{Proof of Lemma~\ref*{lem:ci-overconfidence}}
\AppendixContentsSubsection{app:proof-aliasing-noncollapse}{Proof of Proposition~\ref*{prop:aliasing-noncollapse}}
\AppendixContentsSubsection{app:proof-dealiasing-drift}{Proof of Proposition~\ref*{prop:dealiasing-drift}}
\AppendixContentsSubsection{app:proof-marginal-insufficiency}{Proof of Proposition~\ref*{prop:marginal-insufficiency}}
\AppendixContentsSubsection{app:proof-planner-optimality}{\nameref*{app:proof-planner-optimality}}
\AppendixContentsSection{app:implementation-details}{Computational Complexity}
\AppendixContentsSection{app:notation}{Key Notation}
\AppendixContentsSection{app:exp-supplement}{Supplementary Experimental Results}
\AppendixContentsSubsection{app:exp-query-stop}{\nameref*{app:exp-query-stop}}
\AppendixContentsSubsection{app:exp-robustness}{\nameref*{app:exp-robustness}}
\AppendixContentsSubsection{app:exp-fever-details}{\nameref*{app:exp-fever-details}}
\AppendixContentsSubsection{app:exp-health-details}{\nameref*{app:exp-health-details}}
\AppendixContentsSubsection{app:exp-external-details}{\nameref*{app:exp-external-details}}
\end{tabularx}\par
\endgroup
\medskip

\section{Proofs of Theoretical Properties}
\label{app:proofs}

\subsection{Proof of Lemma~\ref{lem:ci-overconfidence}}
\label{app:proof-ci-overconfidence}
\textit{Proof.}\quad
Under the CI likelihood that ignores shared bias, the one-step increment in the correctness log-odds is
\[
\Delta\lambda_t^{\mathrm{CI}}
=\log\frac{\sigma(e_t w_{\ActI})}{\sigma(-e_t w_{\ActI})}
=e_t w_{\ActI}.
\]
Under the true configuration $(z,u)=(0,1)$, with the task context and held label fixed,
Assumptions~\ref{asmp:static-xi} and~\ref{asmp:cond-indep-z-xi} make repeated \ActInspect
observations independent and identically distributed, with
\[
\Pr_{\mathrm{true}}(e_t=+1)=\sigma(-w_{\ActI}+d_{\ActI}).
\]
Therefore,
\[
\mathbb{E}[\Delta\lambda_t^{\mathrm{CI}}]
=w_{\ActI}\bigl[2\sigma(d_{\ActI}-w_{\ActI})-1\bigr].
\]
When $d_{\ActI}>w_{\ActI}$, this expectation is strictly positive. Because the increments are
bounded and i.i.d., the strong law of large numbers gives
$(\lambda_t^{\mathrm{CI}}-\lambda_0^{\mathrm{CI}})/t\to\mathbb E[\Delta\lambda_t^{\mathrm{CI}}]>0$
a.s. Hence, $\lambda_t^{\mathrm{CI}}\to+\infty$ a.s. and
$b_t^{\mathrm{CI}}=\sigma(\lambda_t^{\mathrm{CI}})\to1$ a.s., even though the true label state is
$z=0$.
\hfill$\square$

\subsection{Proof of Proposition~\ref{prop:aliasing-noncollapse}}
\label{app:proof-aliasing-noncollapse}
\textit{Proof.}\quad
We first show that the posterior probability ratio of the two aliased explanations remains constant,
then show that the other two explanations lose all posterior mass. These two steps determine the
limiting correctness belief. Because $r_{t,j}=0$ for every $j\neq k$, the \ActInspect likelihood
depends only on $(z,\xi_k)$. Summing~\eqref{eq:joint-bayes-update} over all remaining bias
coordinates shows that~\eqref{eq:single-regime-marginal} follows an exact four-state Bayesian
update. Let
\[
m_{\ActI}(z,u)=(2z-1)w_{\ActI}+u d_{\ActI}.
\]
When $d_{\ActI}=2w_{\ActI}$, we have $m_{\ActI}(1,0)=m_{\ActI}(0,1)=w_{\ActI}$, so the two states
have the same likelihood for every $e$. Their common factor cancels in the Bayesian update, which
immediately yields~\eqref{eq:within-class-ratio}.

The observation law under the true explanation $(0,1)$ is $\mathrm{Bern}(\sigma(w_{\ActI}))$. The
other two explanations satisfy $m_{\ActI}(0,0)=-w_{\ActI}$ and $m_{\ActI}(1,1)=3w_{\ActI}$, so their
observation laws differ from the true law. Their posterior log-probability ratios relative to the
true explanation are random walks with bounded i.i.d. increments whose respective means are
\[
-D_{\mathrm{KL}}\!\left(\mathrm{Bern}(\sigma(w_{\ActI}))\middle\|\mathrm{Bern}(\sigma(-w_{\ActI}))\right)<0
\]
and
\[
-D_{\mathrm{KL}}\!\left(\mathrm{Bern}(\sigma(w_{\ActI}))\middle\|\mathrm{Bern}(\sigma(3w_{\ActI}))\right)<0.
\]
By the strong law of large numbers, both log-probability ratios tend to $-\infty$, so
$P_{t,k}(0,0),P_{t,k}(1,1)\to0$ a.s. The posterior mass therefore concentrates on the two aliased
explanations, while their probability ratio remains equal to its initial value.
By~\eqref{eq:joint-prior},
\[
P_{0,k}(1,0)=b_0(1-\rho_{0,k}),\qquad
P_{0,k}(0,1)=(1-b_0)\rho_{0,k},
\]
Substitution into $b_t=P_{t,k}(1,0)+P_{t,k}(1,1)$ yields~\eqref{eq:prop3-pi}.
\hfill$\square$

\begin{remark}[Discriminative information near aliasing]
To examine how much information \ActInspect provides away from exact aliasing, retain the
single-bias setting above, fix $w_{\ActI}$ and the held label, and let
$d_{\ActI}=2w_{\ActI}+\epsilon$. The true explanation remains $(0,1)$; the planner uses the correct
parameters, and both explanations under comparison have positive prior probability. Let
$p=\sigma(w_{\ActI})$ and $p_\epsilon=\sigma(w_{\ActI}+\epsilon)$. The \ActInspect observation laws
under the two explanations are then $\mathrm{Bern}(p_\epsilon)$ and $\mathrm{Bern}(p)$,
respectively. With $\Lambda_t=\log[P_{t,k}(0,1)/P_{t,k}(1,0)]$, the Bayesian update gives
\begin{equation}
\mathbb E_{p_\epsilon}[\Lambda_{t+1}-\Lambda_t]
=D_{\mathrm{KL}}\!\left(\mathrm{Bern}(p_\epsilon)\middle\|\mathrm{Bern}(p)\right).
\label{eq:near-alias-kl}
\end{equation}
Let $A(\theta)=\log(1+\exp\theta)$, whose derivatives satisfy $A'(\theta)=\sigma(\theta)$ and
$A''(\theta)=\sigma(\theta)[1-\sigma(\theta)]$. Expressing the Bernoulli KL divergence through $A$
and applying a Taylor expansion, as $\epsilon\to0$,
\begin{equation}
\begin{aligned}
D_{\mathrm{KL}}
&=\epsilon A'(w_{\ActI}+\epsilon)
 -A(w_{\ActI}+\epsilon)+A(w_{\ActI})\\
&=\tfrac12 p(1-p)\epsilon^2+O(|\epsilon|^3).
\end{aligned}
\label{eq:near-alias-expansion}
\end{equation}
If \ActInspect is executed $T$ consecutive times without switching the held label, the expected
increment in the posterior log-ratio is $T$ times~\eqref{eq:near-alias-kl}. Near the aliasing point,
the discriminative information supplied by each \ActInspect is therefore second order, and a finite
number of inspections may still fail to separate the two explanations.
\end{remark}

\subsection{Proof of Proposition~\ref{prop:dealiasing-drift}}
\label{app:proof-dealiasing-drift}
\textit{Proof.}\quad
We first show that the two explanations aliased under \ActInspect have different observation
distributions under \ActRelabel. The exact aliasing relation $d_{\ActI}=2w_{\ActI}$ gives
$r=2w_{\ActI}/\varphi_{\ActI,k}$. If \ActRelabel also failed to distinguish the bias explanation
$(0,1)$ from the correct explanation $(1,0)$, then
\(-w_{\ActR}+d_{\ActR}=w_{\ActR}\), or equivalently $d_{\ActR}=2w_{\ActR}$. Substituting $d_{\ActR}=\varphi_{\ActR,k}r$ makes this condition equivalent to
$w_{\ActI}\varphi_{\ActR,k}=w_{\ActR}\varphi_{\ActI,k}$, contradicting the assumed inequality $w_{\ActI}\varphi_{\ActR,k}\neq w_{\ActR}\varphi_{\ActI,k}$. The two \ActRelabel observation laws must therefore differ.

Before the held label is switched, the Bernoulli observation distributions under the bias and
correct explanations are $p_{\ActR}^{\mathrm{bias}}$ and $p_{\ActR}^{\mathrm{correct}}$,
respectively. The Bayesian update gives
\[
\widetilde\Lambda_{t+1}-\Lambda_t
=\log\frac{p_{\ActR}^{\mathrm{bias}}(e_t)}
{p_{\ActR}^{\mathrm{correct}}(e_t)}.
\]
Taking expectation under the true observation distribution $p_{\ActR}^{\mathrm{bias}}$
yields~\eqref{eq:prop3-drift}. Because the two distributions differ, Gibbs' inequality makes the KL
divergence strictly positive. This increment is computed before \ActRelabel switches the held label
and reindexes the correctness state.
\hfill$\square$

\subsection{Proof of Proposition~\ref{prop:marginal-insufficiency}}
\label{app:proof-marginal-insufficiency}
\textit{Proof.}\quad
We construct two joint beliefs with identical marginals but different optimal actions, showing that
the correctness and bias marginals alone do not determine the optimal action. Consider two planning
states with $K=1$ and remaining horizon $H=2$. They share the same held label, cue direction $+1$,
and accumulated cost $C=0$. Set
$w_{\ActI}=1$, $w_{\ActR}=1.2$, $w_{\ActD}=1.8$,
$d_{\ActI}=2$, $d_{\ActR}=0.8$, $d_{\ActD}=0.1$,
$c_{\ActI}=1$, $c_{\ActR}=3$, $c_{\ActD}=8$, $c_{\mathrm{err}}=20$, $c_{\ActU}=10$, and $C_{\max}=8$.
To construct a counterexample, consider the following family of joint beliefs:
\begin{equation}
\begin{aligned}
P^{(q)}(0,0)=P^{(q)}(1,1)&=q,\\
P^{(q)}(0,1)=P^{(q)}(1,0)&=0.5-q.
\end{aligned}
\label{eq:interior-coupling-family}
\end{equation}
For any $q\in(0,0.5)$, direct summation gives
$b=P^{(q)}(1,0)+P^{(q)}(1,1)=0.5$ and
$\rho=P^{(q)}(0,1)+P^{(q)}(1,1)=0.5$.
Thus, varying $q$ changes the dependence between correctness and bias while leaving both marginals
unchanged.
Take $q=0.01$ and $q=0.49$. Let $S_q$ denote the corresponding planning state, and define
$p_e^a=\Pr(e\mid S_q,a;\mathbf x)$ and $S_{q,e}^a=\tau(S_q,a,e)$. For a continuing action, the
two-step recursion is
\[
Q_2(S_q,a)=c_a+\sum_{e\in\{-1,+1\}}p_e^a
\min_{a'\in\mathcal A^{\mathrm{feas}}(S_{q,e}^a)}
Q_1(S_{q,e}^a,a').
\]
The feasible set in this recursion is~\eqref{eq:A-feas} evaluated at the successor state. At each
successor, $Q_1$ is computed by~\eqref{eq:Q-cont} and evaluates either immediate termination or one
continuing action followed by termination. If \ActRelabel returns a different label, the successor
reverses both the correctness coordinate and the cue direction.

Substituting $q=0.01$ and $q=0.49$ into~\eqref{eq:unified-obs}, the posterior update, and the
two-step Bellman recursion gives the following approximate expected losses:
\begin{center}
\footnotesize
\setlength{\tabcolsep}{3pt}
\begin{tabular}{@{}crrrrr@{}}
\toprule
$q$ & \ActAccept & \ActInspect & \ActRelabel & \ActArbitrate & \ActUnresolved \\
\midrule
0.01 & 10.000 & 10.271 & 9.271 & 10.958 & 10.000 \\
0.49 & 10.000 & 5.947 & 6.505 & 10.724 & 10.000 \\
\bottomrule
\end{tabular}
\end{center}
For example, at $q=0.01$, the two \ActRelabel branch probabilities are approximately $0.584343$ and
$0.415657$, and both successors select \ActAccept. Their remaining losses give
\[
\begin{aligned}
Q_2(S_{0.01},\ActR)
&\approx3+0.584343(6.809625)\\
&\quad+0.415657(5.514877)
\approx9.271454.
\end{aligned}
\]
The unique optimal action is \ActRelabel at $q=0.01$ and \ActInspect at $q=0.49$. All four state
probabilities in both joint beliefs are at least $0.01$, so both beliefs lie strictly inside the
probability simplex. They have identical marginals but different unique optimal actions, proving
that $(b,\rho)$ is generally insufficient to determine the optimal action.

For fixed parameters, costs, and budget, finite-horizon action values vary continuously with the
joint belief. The optimal action in each case has a strictly positive loss margin and therefore
remains optimal in a sufficiently small neighborhood of the corresponding belief.
\hfill$\square$

\subsection{Finite-Horizon Planning Guarantees}
\label{app:proof-planner-optimality}
\begin{proposition}[Optimality and monotonicity with planning depth]
\label{prop:finite-horizon-optimality}
Under the observation model in~\eqref{eq:unified-obs}, fixed task context $\mathbf x$, known model
parameters, planning state in~\eqref{eq:planning-state}, and feasible set in~\eqref{eq:A-feas}, the
following statements hold:
\begin{enumerate}
    \item The backward induction in~\eqref{eq:bellman-V}--\eqref{eq:Q-cont} solves the remaining $\bm{\mathcal P}_0$ problem from any state $S$.
    \item For any $0\le H<h(S)$,
    \begin{equation}
    V_{H+1}(S)\le V_H(S).
    \label{eq:truncation-monotonicity}
    \end{equation}
    Consequently, for $h(S)\ge1$, we have $V_{h(S)}(S)\le V_1(S)$.
\end{enumerate}
\end{proposition}
\textit{Proof.}\quad
We verify optimality by backward induction, then establish monotonicity by comparing the feasible
policies at successive planning depths. Given \(S\), the fixed context \(\mathbf x\), and the model
parameters, the observation probabilities in~\eqref{eq:predictive-e}, the successors defined by
\(\tau\), and the terminal losses in~\eqref{eq:term-loss} are determined. No additional history is
needed to evaluate subsequent decisions.

At depth \(0\), no continuing action is allowed within the planning horizon, so
\(V_0(S)=J_{\mathrm{term}}(S)\). Suppose \(V_{h-1}\) gives the minimum expected remaining loss at
every successor. At depth \(h\), the first action of any feasible policy is either a terminal action
with loss \(L_a(S)\), or a continuing action that incurs \(c_a\) and then averages \(V_{h-1}\) over
the branches according to~\eqref{eq:predictive-e}. Minimizing over these first actions
gives~\eqref{eq:bellman-V}. Taking \(h=h(S)\) proves the first statement.

For the second statement, every feasible policy using at most \(H\) continuing actions remains
feasible when up to \(H+1\) such actions are allowed. Minimizing over this larger policy class
gives~\eqref{eq:truncation-monotonicity}. For any feasible continuing action \(a\in\{\ActI,\ActR\}\)
and \(1\le h\le h(S)\), substituting \(V_{h-1}(S')\le V_0(S')\) at each successor
into~\eqref{eq:Q-cont} also gives \(Q_h(S,a)\le Q_1(S,a)\).
\hfill$\square$

\nobalance
\section{Computational Complexity}
\label{app:implementation-details}

\subsection{Exact Planning Cost}
\label{app:planning-complexity}

The cost of exact planning depends on the size of the joint-belief table and the number of branches
explored. Each node processes $2^{K+1}$ joint-belief entries, and direct computation of the bias
shift in~\eqref{eq:Delta-def} requires $\mathcal O(K)$ work per entry. From $S_t$, \ActInspect and
\ActRelabel each produce two possible observation branches, giving at most four children per node.
The tree has depth at most $h=T_{\max}-t$. Writing $\mathcal A_{\mathrm{cont}}=\{\ActI,\ActR\}$, the
worst-case planning cost is
\[
\mathcal O\!\left(
\sum_{d=0}^{h}
\bigl(2|\mathcal A_{\mathrm{cont}}|\bigr)^d
\cdot K\,2^{K+1}
\right).
\]
The budget constraint prunes infeasible branches, but the worst-case bound remains exponential in
both the number of bias types and the planning depth. The current exact recursion is therefore
suited to small $K$ and short horizons; larger problems require further pruning or approximate
methods.

\section{Key Notation}
\label{app:notation}
Table~\ref{tab:notations} summarizes the main notation.
\begin{table}[H]
\centering
\caption{Key notation}
\label{tab:notations}
\footnotesize
\renewcommand{\arraystretch}{1.02}
\setlength{\tabcolsep}{3pt}
\begin{tabular}{@{}p{0.29\columnwidth}p{0.67\columnwidth}@{}}
\toprule
Symbol & Meaning\\
\midrule
$\mathcal Y,y^*$ & Binary label space and true label.\\
$\hat y_{\mathrm{init}},\hat y_t$ & Initial and currently held labels.\\
$\mathbf x,z_t$ & Task attributes; held-label correctness.\\
$K,\xi_k,\xi$ & Number of bias types, activity indicator, and bias vector.\\
$P_0,P_t$ & Initial and current joint beliefs.\\
$b_t,\lambda_t$ & Correctness probability and log-odds.\\
$\rho_{t,k},\bm\rho_t$ & Bias-activity probability and its vector.\\
$s_{t,k},m_{t,k},r_{t,k}$ & Cue direction, intensity, and their product.\\
$w_a$ & Baseline diagnostic strength.\\
$\varphi_{a,k},\boldsymbol\Phi$ & Action--bias coupling and its matrix.\\
$\Delta_t(a,\xi),d_a$ & Total bias shift; single-bias shift.\\
$e_t,\mathcal L_t$ & Binary observation and likelihood.\\
$P_{t,k},\Lambda_t$ & Joint belief of $(z,\xi_k)$; log posterior odds of $(0,1)$ versus $(1,0)$.\\
\midrule
$\mathcal A,\widetilde{\mathcal A}$ & QC actions; extended set including \ActUnresolved{}.\\
$\ActA,\ActI,\ActR,\ActD,\ActU$ & \ActAccept{}, \ActInspect{}, \ActRelabel{}, \ActArbitrate{}, \ActUnresolved{}.\\
$\mathcal A_t^{\mathrm{feas}}$ & Feasible actions at step $t$.\\
$\mathcal A_{\mathrm{term}}$ & Terminal actions $\{\ActA,\ActD,\ActU\}$.\\
$c_a,c_{\mathrm{err}},c_{\ActU}$ & Action, output-error, and \ActUnresolved{} costs.\\
$C_t,C_{\max}$ & Accumulated acquisition cost and budget.\\
$T_{\max},T_{\mathrm{stop}}$ & Acquisition-step limit and stopping time.\\
\midrule
$\pi,S_t$ & Policy; state $(P_t,\hat y_t,C_t,t)$.\\
$V_h(S),Q_h(S,a)$ & Minimum expected remaining loss; expected loss with first action $a$ and depth $h$.\\
$L_a,J_{\mathrm{term}}$ & Expected terminal loss; its minimum over terminal actions.\\
$\tau$ & State transition.\\
$\alpha_{\ActD}(S;\mathbf x)$ & Arbitration correctness probability at state $S$.\\
$\bm{\mathcal P}_0$ & Constrained expected-loss minimization.\\
\bottomrule
\end{tabular}
\end{table}

\section{Supplementary Experimental Results}
\label{app:exp-supplement}

This appendix provides the protocols and supplementary evidence for Section~\ref{sec:experiments}.

\subsection{Controlled Grids, Planning, and Reference Baselines}
\label{app:exp-query-stop}
\label{app:exp-routing}

This subsection supplements Section~\ref{subsec:exp-simulation} with parameter configurations and
sequential-query comparisons. The experiments vary bias coupling or planning depth to distinguish
how the joint posterior changes evidence evaluation and how these changes affect decision loss.

\smallskip\noindent\textbf{Main comparison and grid definitions.}
The main comparison uses prespecified simulation configurations. For $K=2$, both bias variables have
activation probability $\Pr(\xi_k=1)=0.25$, are sampled independently at the beginning of each
episode, and then remain fixed. The initial cue directions are $(+1,-1)$, and the bias-coupling coefficients
$(\varphi_{a,1},\varphi_{a,2})$ for \ActInspect{}, \ActRelabel{}, and \ActArbitrate{} are $(2.0,0.8)$, $(0.8,2.4)$, and $(0.1,0.1)$,
respectively. \ActInspect{} and \ActRelabel{} have different sensitivities to the two biases,
allowing a switch between these actions to distinguish latent explanations of the same observation
history. The $K=3$ configuration adds a third bias with activation probability 0.20, direction $+1$, and
coupling coefficients $\varphi_{\ActI,3}=\varphi_{\ActR,3}=1.2$ and $\varphi_{\ActD,3}=0.1$,
examining an additional bias that affects both \ActInspect{} and \ActRelabel{}.

Unless stated otherwise, the diagnostic strengths are $(w_{\ActI},w_{\ActR},w_{\ActD})=(1,1.2,1.8)$,
action costs are $(c_{\ActI},c_{\ActR},c_{\ActD})=(1,3,8)$, the acquisition budget is $C_{\max}=8$,
and the maximum number of continuing QC steps is $T_{\max}=6$. We use $b_0\in\{0.50,0.60,0.70\}$.
Costs are normalized by $c_{\ActI}=1$; \textit{Original} uses erroneous-output and \ActUnresolved{}
penalties $(c_{\mathrm{err}},c_{\ActU})=(20,10)$, while \textit{Risk-sensitive} uses $(80,20)$. All
expected losses are computed by full path enumeration.

The single-bias scan uses activation probability 0.25, with no bias coupling to \ActArbitrate{}. Its
coupling values are $d_{\ActI}\in\{1.2,1.6,2.0,2.4,2.8\}$ and $d_{\ActR}\in\{0,0.6,1.2,1.8,2.4\}$.
The exact \ActInspect{} alias point is $d_{\ActI}=2.0$, and the corresponding \ActRelabel{} alias
point is $d_{\ActR}=2.4$. The grid therefore covers both sides of \ActInspect{} aliasing and varies
\ActRelabel{} from an action unaffected by bias to one that cannot distinguish the corresponding
competing explanations.

\smallskip\noindent\textbf{Grid outcomes.}
Across the 75 single-bias settings, \textit{JABER} reduces total loss relative to
\textit{Marginal-Projection} in 33 and 72 settings under \textit{Original} and
\textit{Risk-sensitive} weights, respectively; the methods coincide without shared bias.

Table~\ref{tab:exp-loss-decomposition} decomposes the JABER--Marginal-Projection loss differences in
the main two- and three-bias comparisons. Under \textit{Original} weights, savings in action costs
and unresolved-outcome losses outweigh the higher error losses. For $K=2$, the action-cost savings
and error-loss increase nearly cancel, so the reduction in unresolved-outcome loss accounts for most
of the net gain. Under \textit{Risk-sensitive} weights, lower error losses dominate; action costs
increase slightly for $K=2$ but decrease for $K=3$.

\begin{table}[t]
\centering
\caption{Decomposition of the expected QC loss difference between \textit{JABER} and
\textit{Marginal-Projection}. Entries are \textit{JABER} minus \textit{Marginal-Projection},
averaged equally over $b_0\in\{0.50,0.60,0.70\}$. $\Delta L$ is the total loss difference;
$\Delta L_{\rm act}$, $\Delta L_{\rm err}$, and $\Delta L_{\ActU}$ are its action-cost,
erroneous-output, and unresolved-outcome components. $K$ denotes the number of shared-bias types.
Negative values indicate lower loss for \textit{JABER}.}
\label{tab:exp-loss-decomposition}
\scriptsize
\setlength{\tabcolsep}{3.2pt}
\begin{tabular}{@{}clrrrr@{}}
\toprule
$K$ & Weights & $\Delta L\downarrow$ & $\Delta L_{\rm act}\downarrow$ & $\Delta L_{\rm err}\downarrow$ & $\Delta L_{\ActU}\downarrow$ \\
\midrule
2 & \textit{Original} & $-0.214$ & $-1.267$ & $+1.215$ & $-0.161$ \\
3 & \textit{Original} & $-0.156$ & $-0.464$ & $+0.364$ & $-0.055$ \\
2 & \textit{Risk-sensitive} & $-0.842$ & $+0.038$ & $-0.880$ & $0$ \\
3 & \textit{Risk-sensitive} & $-1.192$ & $-0.103$ & $-1.090$ & $0$ \\
\bottomrule
\end{tabular}
\end{table}

\smallskip\noindent\textbf{Planning depth.}
The main-text horizon scan changes the number of available decision steps. The \textit{One-step
lookahead} comparison here instead keeps the task horizon fixed and retains joint updating, but
truncates planning depth to $h=1$. The two comparisons test whether future decision opportunities
are available and whether the planner uses them, respectively. In the five-action single-bias
setting, the false-accept differences between full planning and \textit{One-step lookahead} are
$-3.51$ and $+5.11$ percentage points under \textit{Original} and \textit{Risk-sensitive} weights.
Planning gains should therefore be evaluated through total loss rather than false accepts alone.

\smallskip\noindent\textbf{Matched-action \textit{Query/Stop} comparisons.}
To compare sequential-query policies, these experiments use a separate \textit{Query/Stop} task: a
method can acquire the next judgment or stop and decide whether the held label is correct, without
\ActRelabel{}, \ActArbitrate{}, or \ActUnresolved{}. All methods receive the same initial beliefs
and same-source judgments, pay query cost 1, and acquire at most six judgments. Error losses are 20
and 80 under \textit{Original} and \textit{Risk-sensitive} weights. The alias configuration uses
$w_{\ActI}=1$, bias prior 0.3, and coupling 2, with a no-bias control retained.

\textit{CLARA-based Query/Stop} uses the homogeneous-reviewer special case of CLARA to estimate
correctness under conditional independence and applies the same finite-horizon risk recursion to
decide whether to query or stop~\cite{nguyen2020clara}. \textit{Raykar-style} compares the next
judgment's expected information gain under a conditionally independent posterior with its query
cost, scaling information gain by $c_{\mathrm{err}}/2$~\cite{raykar2014sequential}. Fixed-$k$
acquires $k\in\{1,3,5\}$ judgments and then outputs a MAP decision from the conditionally
independent posterior. These are task-matched adaptations that compare the corresponding posterior
or query mechanisms. Table~\ref{tab:exp-reference-baselines} reports equal-weight averages over the
three initial beliefs. \textit{JABER} has lower total loss than the listed baselines, although
\textit{Fixed-3} and \textit{Fixed-5} have fewer false accepts under \textit{Original} weights.

\begin{table}[H]
\centering
\caption{Single-bias alias-stress comparisons, averaged over $b_0\in\{0.50,0.60,0.70\}$ by exact
enumeration. Positive relative loss gain and negative false-accept difference ($\Delta$FA, in
percentage points) favor \textit{JABER}. All rows compare baselines in the matched
\textit{Query/Stop} task.}
\label{tab:exp-reference-baselines}
\scriptsize
\setlength{\tabcolsep}{2.7pt}
\begin{tabular}{@{}lrrrr@{}}
\toprule
& \multicolumn{2}{c}{\textit{Original}} & \multicolumn{2}{c}{\textit{Risk-sensitive}} \\
\cmidrule(lr){2-3}\cmidrule(l){4-5}
Baseline & Loss gain $\uparrow$ & $\Delta$ FA $\downarrow$ & Loss gain $\uparrow$ & $\Delta$ FA $\downarrow$ \\
\midrule
\makecell[l]{\textit{CLARA-based}\\\textit{Query/Stop}} & 2.15\% & $-2.09$ pp & 1.58\% & $-1.96$ pp \\
\textit{Raykar-style} & 5.26\% & $-0.95$ pp & 7.62\% & $-3.05$ pp \\
\textit{Fixed-1} & 5.17\% & $-0.14$ pp & 22.72\% & $-2.45$ pp \\
\textit{Fixed-3} & 15.29\% & $+1.32$ pp & 12.35\% & $-0.99$ pp \\
\textit{Fixed-5} & 28.29\% & $+2.18$ pp & 9.88\% & $-0.13$ pp \\
\bottomrule
\end{tabular}
\end{table}

\subsection{Misspecification Robustness}
\label{app:exp-robustness}

The following checks examine sensitivity to parameter perturbations, an omitted bias, and
observation-link misspecification.

\smallskip\noindent\textbf{Parameter perturbations and omitted bias.}
We fix the $K=3$ generative model and perturb the planner's coupling coefficients by $\pm20\%$ or
its bias priors by $\pm40\%$, yielding 24 misspecified settings. Against
\textit{Marginal-Projection} with the same misspecified parameters, \textit{JABER} is never worse
and is strictly better in 21 settings. To quantify the cost of misspecification itself, we also
compare misspecified \textit{JABER} with \textit{JABER} using the true parameters: the median loss
increase across the 24 settings is 0.17\%, and the maximum is 4.19\%. In a separate omitted-variable
check, the generative process retains a third bias affecting both \ActInspect{} and \ActRelabel{}
that the planner does not model; \textit{JABER} still outperforms \textit{Marginal-Projection} in
all six settings.

\smallskip\noindent\textbf{Observation-link misspecification.}
To test observation-link misspecification, we vary $\varphi_{\ActI,1}\in\{1.6,2.0,2.4\}$ and
$\varphi_{\ActR,2}\in\{2.0,2.4,2.8\}$, using the three initial beliefs and remaining $K=2$
parameters in Appendix~\ref{app:exp-query-stop}. This gives 27 settings per loss profile.
Observation probabilities follow either a probit or a centered complementary log-log link, while the
planners retain the original sigmoid likelihoods and coefficients. Writing $\eta$ for the logit of
the support observation $e=+1$, the alternative links are
\[
\begin{aligned}
g_{\mathrm P}(\eta)&=\Phi_{\mathrm N}\!\left(\sqrt{\pi/8}\,\eta\right),\\
g_{\mathrm C}(\eta)&=1-\exp\!\left[-\ln 2\,\exp\!\left(\frac{\eta}{2\ln 2}\right)\right],
\end{aligned}
\]
where $\Phi_{\mathrm N}$ is the standard normal cumulative distribution function. For each link $g$,
support and opposition have probabilities $g(\eta)$ and $1-g(\eta)$, respectively. Both links match
the sigmoid's value and slope at zero. Expected losses are evaluated by exact path enumeration under
each generating link.

\begin{table}[H]
\centering
\caption{Mean loss reduction of \textit{JABER} relative to \textit{Marginal-Projection} under
observation-link misspecification.}
\label{tab:exp-link-misspec}
\footnotesize
\begin{tabular}{@{}lrr@{}}
\toprule
Data-generating link & \multicolumn{2}{c}{Mean loss reduction $\uparrow$} \\
\cmidrule(lr){2-3}
& \textit{Original} & \textit{Risk-sensitive} \\
\midrule
Sigmoid (matched control) & 2.51\% & 4.77\% \\
Probit & 2.30\% & 5.81\% \\
Centered complementary log-log & 2.63\% & 5.90\% \\
\bottomrule
\end{tabular}
\end{table}

Reductions in Table~\ref{tab:exp-link-misspec} are averaged equally across the 27 settings. For each
link, \textit{JABER} has lower loss in 25 \textit{Original} settings and ties in two; it has lower
loss in all 27 \textit{Risk-sensitive} settings.

\subsection{\textit{FEVER} Protocol and Supplementary Comparisons}
\label{app:exp-fever-details}

This subsection provides the data, calibration, and evaluation protocol for
Section~\ref{subsec:exp-fever}. Tables~\ref{tab:fever-full-primary} and~\ref{tab:fever-all-costs}
report the complete baseline comparisons and sensitivity to terminal-loss weights.

\smallskip\noindent\textbf{Data and QC actions.}
We construct separate development and evaluation sets to calibrate observation probabilities from
real API judgments and assess the resulting QC policies. To vary the influence of a shared
information source, each of the 16 development claims is paired with two simulated background
accounts supporting opposite conclusions and two review instructions: independently verify the claim
or prioritize the supplied account. Each account/instruction combination provides three same-prompt
\ActInspect{} draws and one \ActRelabel{} response. The account and instruction remain fixed within
an episode, while the planner is not informed which instruction is used. Assigning each of the two
possible initial labels to every combination gives 128 weighted development views. These views reuse
the responses for a given combination to evaluate updating and decisions from different initial
labels; the number of independent development claims remains 16.

After development, we evaluate on 40 new claims with no overlapping claims or annotated evidence
pages. Each claim forms one QC episode, with 32 supplied initial labels correct and eight incorrect.
Gold labels guide material construction, development calibration, and final scoring. During
execution, neither the LLMs nor the planner receive gold labels, and retrieval evidence is not
selected from the gold annotations.

\ActInspect{} uses \textit{Qwen-Flash} (\texttt{qwen-flash-}\allowbreak\texttt{2025-07-28}) at temperature 0.7; \ActRelabel{} uses \textit{Qwen-Plus} (\texttt{qwen-plus-}\allowbreak\texttt{2025-12-01}) at temperature 0; and \ActArbitrate{} uses \textit{DeepSeek} (\texttt{deepseek-}\allowbreak\texttt{v4.1-flash}) at temperature 0. \textit{Cheap Flash Relabel} uses \textit{Qwen-Flash} at temperature 0 with retrieved evidence alone. Retrieval uses the June 2017 Wikipedia snapshot; retrieved evidence can be incomplete or insufficient for a correct judgment. All requests omit the held label and previous judgments, with streaming and thinking disabled. There are no response-quality retries or outcome-selected exclusions. Returned support/refutation verdicts are encoded as support for or opposition to the current held label, and re-encoded after \ActRelabel{} changes that label.

\smallskip\noindent\textbf{Observation model and calibration.}
\textit{JABER} retains the observation model in~\eqref{eq:unified-obs}. Development fitting compares
single-source, source-and-polarity, and graded-source variants, representing one shared source, two
distinct bias directions, and two levels of influence from the same source, respectively.
Cross-validation selects the graded-source model. Both bias cues point toward the label supported by
the background account; one source component is known active, while the other represents latent
additional susceptibility. This component is inferred from responses, and neither review instruction
is used as a supervised target for it. The initial belief factorizes with \(b_0=0.8\) and bias
priors \((1,0.657731)\). The fitted strengths are
\((w_{\ActI},w_{\ActR},w_{\ActD})=(4.146201,4.986115,5.931898)\). The coupling pairs for
\ActInspect{} and \ActRelabel{} are \((0.760424,6.453605)\) and \((0.995547,2.069058)\),
respectively; both \ActArbitrate{} couplings are fixed at zero.

To compare the expected losses of acquiring evidence and terminating QC, the planner needs
probabilities for label correctness and QC judgments. We fit these probabilities on development data
so that the model predicts correctness and also explains the collected response sequences. Let
\(L_z\) be the negative log score for held-label correctness after \ActInspect{} or \ActRelabel{},
\(L_D\) the negative log score for arbitration-label correctness at each decision state, and
\(L_{\mathrm{seq}}\) the negative log score of the response sequence after marginalizing latent
states. The first two scores are averaged over available decision states within each view; the
sequence score is normalized by the number of observed responses. The scores are then averaged over
views with correct and incorrect initial labels weighted 0.20 and 0.05, respectively. These weights
give total weight one per claim and initial correctness 0.8. The fitting objective is
\begin{equation}
\mathcal J(\theta)
=W\!\left[L_z(\theta)+L_D(\theta)
+\lambda L_{\mathrm{seq}}(\theta)\right]
+\gamma\|\theta\|_2^2,
\label{eq:fever-calibration-objective}
\end{equation}
where \(W\) is the sum of view weights, \(\theta\) is the optimization parameter vector, and
\(\gamma=0.1\). The first two terms score the correctness probabilities used in QC decisions, while
the sequence term fits the observed judgments using the same parameters. The weight \(\lambda\)
controls the contribution of sequence fitting, and the final term penalizes large parameter values.
We select \(\lambda\) by three-fold cross-validation, keeping all views from the same claim in one
fold. The candidates \(\{0,0.5,2,8,32\}\) are evaluated using the common validation score
\(L_z+L_D+L_{\mathrm{seq}}\). The selected weight is \(\lambda=32\), after which the model is
refitted on all 16 development claims.

To check whether recalibration, ordinary instance difficulty, or more general response dependence
accounts for the joint model's gains, we also fit \textit{Calibrated CI}, the \textit{ordinary
difficulty model}, and \textit{Full Ising} using the same splits, fitting scores, and
model-selection procedure. The \textit{difficulty model} represents persistent variation in judgment
reliability. \textit{Full Ising} represents all ten pairwise interactions among three \ActInspect{}
responses, \ActRelabel{}, and \ActArbitrate{}. These models use the same sequential planner,
horizon, and resource constraints. \textit{CI Updating} instead removes the bias terms from
\textit{JABER}'s observation model.

\smallskip\noindent\textbf{Evaluation protocol.}
The development grid varies error loss, the \ActUnresolved{}-to-error loss ratio, the acquisition
horizon, and the forecast budget. We select a primary setting with both inspection-only acceptance
and replacement paths, lower development loss than \textit{CI Updating},
\textit{Marginal-Projection}, and \textit{One-step lookahead}, and competitive loss against strong
fixed rules and direct replacement. The selected setting ties with the \textit{best precommitted
sequence} on development data. The error loss is 1,280 times the mean development \ActInspect{} fee,
giving 0.058678 CNY; the \ActUnresolved{} loss is 10\% of this value. These development-selected
utility weights are frozen before testing and are distinct from usage-based API fees.

The forecast fees for \ActInspect{}, \ActRelabel{}, and \ActArbitrate{} are 0.0000458422,
0.0007566063, and 0.0025900000 CNY. Acquisition permits at most three \ActInspect{}/\ActRelabel{}
steps and one \ActRelabel{}, with a forecast budget of one \ActRelabel{} plus two \ActInspect{}
actions, \(C_{\max}=0.0008482906\) CNY. \ActArbitrate{} remains a separately paid terminal action
outside this acquisition budget. Forecast fees determine action feasibility, whereas realized fees
enter the final QC loss. Realized fees are estimated from returned token usage and published rates,
including documented time and cache adjustments; the account invoice was not reconciled. Every path
satisfies the forecast constraint, but realized \ActInspect{}/\ActRelabel{} fees exceed the budget
in 7/40 primary \textit{JABER} episodes, with a maximum excess of 0.0006998594 CNY. All such fees
are retained in the loss. Supplied initial labels and manually constructed background accounts are
laboratory inputs whose preparation costs are unmeasured.

Models, utility settings, materials, and random assignments are frozen before independent evaluation
calls, and each claim's decision trees are committed before its first QC call. \textit{JABER}
executes first on each claim, selecting an action before receiving its fresh API response and
choosing the next action from the updated belief. Other policies use a shared per-claim response
pool for paired replay. Missing responses required by their paths are collected; otherwise, policies
reuse the same response for a given action and, for \ActInspect{}, a given draw index. Payload
checks verify fixed inputs and the absence of decision-history inputs, and the pool covers every
comparison path. This pairing shares responses across policies and does not assess variability under
separate API realizations for each baseline. \ActArbitrate{} uses separate API calls for query
generation and the final verdict, with both fees charged to the action. Each policy pays the full
fee of every action on its path. Reported 95\% intervals use 5,000 claim-paired bootstrap resamples,
with loss differences defined as \textit{JABER} minus the comparator.

\smallskip\noindent\textbf{Supplementary comparisons.}
Table~\ref{tab:fever-full-primary} adds the calibrated and dependence models described above and
further fixed policies to the primary comparisons. \textit{Fixed IIR} prespecifies
\ActInspect{}--\ActInspect{}--\ActRelabel{} with the same terminal options. Global and
context-selected escalation rules are selected by development loss from feasible agreement-based
rules; the context variant also uses the public initial label, background-account direction, and
option order. The primary global rule is \ActRelabel{} followed by \ActArbitrate{} on disagreement
with the initial label. The additional calibrated and dependence models do not achieve lower loss
point estimates than \textit{JABER} in the primary setting, although some paired loss-difference
intervals include zero.

\begin{table*}[t]
\centering
\caption{Complete primary \textit{FEVER} comparison. Fees, losses, and intervals are in \(10^{-3}\)
CNY. E/U denotes erroneous outputs/unresolved cases out of 40. Intervals are for \textit{JABER}
minus the comparator's total QC loss.}
\label{tab:fever-full-primary}
\scriptsize
\setlength{\tabcolsep}{5pt}
\begin{tabularx}{\textwidth}{@{}Xrrrr@{}}
\toprule
Method & E/U & Mean API fee $\downarrow$ & Mean QC loss $\downarrow$ & Paired 95\% interval\\
\midrule
\textit{JABER} & 0/0 & 0.8845 & 0.8845 & --\\
\textit{Marginal-Projection} & 0/0 & 0.9088 & 0.9088 & $[-0.0312,-0.0174]$\\
\textit{CI Updating} & 1/0 & 0.5837 & 2.0507 & $[-4.1551,0.5174]$\\
\textit{One-step lookahead} & 0/0 & 1.0656 & 1.0656 & $[-0.2809,-0.0934]$\\
\textit{Best precommitted sequence} & 0/0 & 0.8845 & 0.8845 & $[0.0000,0.0000]$\\
\textit{Development-selected fixed escalation rule} & 0/0 & 1.2719 & 1.2719 & $[-0.7024,-0.1134]$\\
\textit{Always-Relabel} & 1/0 & 0.6738 & 2.1408 & $[-4.2624,0.3791]$\\
\textit{Cheap Flash Relabel} & 3/0 & 0.1090 & 4.5098 & $[-9.1212,0.5971]$\\
\textit{Calibrated CI} & 1/0 & 0.7271 & 2.1941 & $[-4.2049,0.2343]$\\
\textit{Difficulty model} & 0/0 & 1.2719 & 1.2719 & $[-0.6974,-0.1087]$\\
\textit{Full Ising} & 1/0 & 0.6385 & 2.1055 & $[-4.1061,0.3355]$\\
\textit{Fixed IIR} & 0/0 & 0.9190 & 0.9190 & $[-0.0427,-0.0257]$\\
\textit{Context-selected fixed escalation rule} & 1/0 & 0.9262 & 2.3932 & $[-4.5261,0.2121]$\\
\textit{Always-Accept} & 8/0 & 0.0000 & 11.7356 & $[-18.2646,-3.7788]$\\
\textit{Always-Arbitrate} & 0/0 & 2.6422 & 2.6422 & $[-2.1547,-1.3511]$\\
\bottomrule
\end{tabularx}
\end{table*}

To distinguish posterior representation from acquisition order, allowing
\textit{Marginal-Projection} to choose its own \textit{best precommitted sequence} gives mean loss
0.00090884 CNY. Requiring it to use the sequence selected by the joint model instead gives
0.00088455 CNY and reproduces \textit{JABER}'s paths and terminal outputs on every claim. This
decomposition locates the observed joint-representation gain in acquisition-plan evaluation and
selection. These claims show no additional gain from changing acquisition order midway or from
terminal decisions after the same sequence.

We vary terminal penalties to test whether the primary ranking persists under different risk
tradeoffs. All five error-loss/\ActUnresolved{}-fraction pairs are frozen before independent calls:
\((0.058678,0.10)\), \((0.029339,0.10)\), \((0.117356,0.10)\), \((0.010000,0.25)\), and
\((0.058678,0.50)\). The horizon, forecast budget, and response-level API fees are unchanged. The
lower-error setting also changes the \ActUnresolved{} fraction, so its result reflects both changes
rather than error loss alone. Under the higher-\ActUnresolved{} setting, all 15 methods have the
same mean total QC losses and E/U counts as under the primary setting in
Table~\ref{tab:fever-full-primary}. Table~\ref{tab:fever-all-costs} reports the remaining three
settings. In the lower-error setting, \textit{JABER} has 19.33\% higher loss than
\textit{Marginal-Projection}; with doubled error loss, it has 6.43\% higher loss than the
\textit{best precommitted sequence}. These comparisons show that changing terminal penalties can
change method rankings.

\begin{table*}[t]
\centering
\caption{\textit{FEVER} sensitivity to terminal-loss weights. Each cell reports mean total QC loss
in \(10^{-3}\) CNY [E/U counts out of 40]. The half-error, double-error, and lower-error settings
use the loss pairs specified in the text; API fees are unchanged. Primary results are reported in
Table~\ref{tab:fever-full-primary}.}
\label{tab:fever-all-costs}
\scriptsize
\setlength{\tabcolsep}{4pt}
\begin{tabularx}{\textwidth}{@{}Xrrr@{}}
\toprule
& \multicolumn{3}{c}{Mean total QC loss $\downarrow$ [E/U counts]}\\
\cmidrule(lr){2-4}
Method & \textit{Half error} & \textit{Double error} & \textit{Lower error}\\
\midrule
\textit{JABER} & 0.8845 [0/0] & 0.9590 [0/0] & 0.7506 [1/0]\\
\textit{Marginal-Projection} & 0.9088 [0/0] & 0.9668 [0/0] & 0.6290 [1/1]\\
\textit{CI Updating} & 1.3172 [1/0] & 3.5176 [1/0] & 0.7154 [1/3]\\
\textit{One-step lookahead} & 1.0656 [0/0] & 1.9720 [0/0] & 0.9348 [1/0]\\
\textit{Best precommitted sequence} & 0.8845 [0/0] & 0.9011 [0/0] & 0.7506 [1/0]\\
\textit{Development-selected fixed escalation rule} & 1.2719 [0/0] & 1.2719 [0/0] & 0.9238 [1/0]\\
\textit{Always-Relabel} & 1.4073 [1/0] & 3.6077 [1/0] & 0.9238 [1/0]\\
\textit{Cheap Flash Relabel} & 2.3094 [3/0] & 8.9107 [3/0] & 0.8590 [3/0]\\
\textit{Calibrated CI} & 1.3802 [1/3] & 3.6689 [1/0] & 0.8566 [1/3]\\
\textit{Difficulty model} & 1.3339 [0/9] & 1.2719 [0/0] & 0.9238 [1/0]\\
\textit{Full Ising} & 1.3730 [1/2] & 3.9489 [1/0] & 0.8678 [1/2]\\
\textit{Fixed IIR} & 0.9190 [0/0] & 0.9769 [0/0] & 0.8574 [1/1]\\
\textit{Context-selected fixed escalation rule} & 1.6597 [1/0] & 3.8601 [1/0] & 0.7352 [1/0]\\
\textit{Always-Accept} & 5.8678 [8/0] & 23.4712 [8/0] & 2.0000 [8/0]\\
\textit{Always-Arbitrate} & 2.6422 [0/0] & 2.6422 [0/0] & 2.6422 [0/0]\\
\bottomrule
\end{tabularx}
\end{table*}

\subsection{\textit{HealthContradict} Calibration and Supplementary Checks}
\label{app:exp-health-details}

This subsection documents the reference-assisted cost scan and supplementary both-passage replay
used to examine how replacement cost and estimated reliability affect terminal QC decisions.

\smallskip\noindent\textbf{Protocol and calibration.}
\textit{HealthContradict} contains 920 expert-verified health-related YES/NO question--answer
samples, each with two reference passages supporting opposite conclusions. For each selected sample,
we construct two episodes, one under each passage condition. An initial judgment provides the held
label, and four subsequent \ActInspect{} judgments reread the same passage under distinct
reviewer-role prompts. Blind \ActRelabel{} receives the passage selected by the experimenter using
the gold label, without seeing the held label or judgment history. We use 20 gold-NO samples for
calibration and 40 for primary evaluation, yielding 80 evaluation episodes. The calibration and
primary evaluation records are disjoint, although nine underlying questions overlap. All API
judgments are obtained using \texttt{qwen-flash-}\allowbreak\texttt{2025-07-28} at temperature 0.
Unless otherwise specified, policy comparisons use \textit{Risk-sensitive} weights
\((c_{\mathrm{err}},c_{\ActU},c_{\ActI})=(80,20,1)\).

The likelihoods are fitted as discrete probability tables rather than using the sigmoid
parameterization in~\eqref{eq:unified-obs}. For each \ActInspect{} step, we count whether its
response supports or opposes each candidate label, using whether the passage supports the gold
answer to distinguish the two bias states. If $k$ of $n$ calibration records in a correctness/bias
cell support the label, the support probability is estimated as $(k+1)/(n+2)$. The four
step-specific tables are fitted separately. Initial-label accuracy is estimated from 40 judgments
across the two passage conditions, and replacement accuracy from 20 judgments using gold-supporting
passages, both with add-one smoothing. The bias prior is fixed at 0.5, corresponding to the paired
passage conditions. All probabilities are frozen before evaluation. Blind \ActRelabel{} is terminal
in this protocol: its answer is output directly, with expected loss
$c_{\ActR}+c_{\mathrm{err}}(1-\alpha_{\ActR})$, where $\alpha_{\ActR}$ is its calibrated accuracy.
It does not create a further posterior-update step. The other actions are \ActAccept{},
\ActInspect{}, and \ActUnresolved{}; \ActArbitrate{} is unavailable.

\smallskip\noindent\textbf{Reference-assisted cost sensitivity.}
We examine how the cost of reference-assisted \ActRelabel{} changes policy choices. The primary cost
point is selected on calibration data by examining integer \ActRelabel{} costs in ascending order.
We choose $c_{\ActR}=4$, the first value at which \textit{JABER} has lower total loss than both
updating ablations without a higher erroneous-output rate, and freeze this choice before evaluation.
At this cost, \textit{JABER}, \textit{Marginal-Projection}, and \textit{CI Updating} all select
immediate \ActRelabel{} on evaluation, yielding mean loss 10.00 and an erroneous-output rate of
7.5\%. Reference-assisted \ActRelabel{} achieves 92.5\% evaluation accuracy. The measured mean
\ActRelabel{}-to-\ActInspect{} API fee ratio is 1.013.\footnote{Alibaba Cloud Model Studio pricing
page: \url{https://help.aliyun.com/zh/model-studio/model-pricing}; calculated using prices recorded
on August 26, 2026 and saved token usage for 800 \ActInspect{} calls and 100 Blind \ActRelabel{}
calls. Cost accounting also includes responses from 40 additional gold-YES samples.} Near this
ratio, the three planners also select immediate \ActRelabel{}.

\begin{figure}[t]
\centering
\includegraphics[width=\columnwidth]{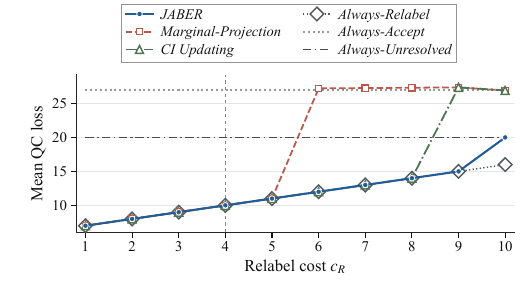}
\caption{\textit{HealthContradict} reference-assisted \ActRelabel{} cost sensitivity. The horizontal
axis gives the \ActRelabel{} cost; the vertical axis gives mean total QC loss over 40 gold-NO
samples with two episodes per sample, with lower values indicating better performance.
\textit{Risk-sensitive} weights are used, with \ActInspect{} cost normalized to 1. The vertical line
marks the calibration-selected \(c_{\ActR}=4\).}
\label{fig:health-cost-appendix}
\end{figure}

Figure~\ref{fig:health-cost-appendix} reports every integer $c_{\ActR}$ from 1 to 10. \textit{JABER}
matches \textit{Always-Relabel} throughout $c_{\ActR}=1$--$9$, giving no additional loss benefit
over direct replacement in this range. At $c_{\ActR}=10$, \textit{JABER} chooses \ActUnresolved{}
with loss 20: its calibrated replacement accuracy implies loss 20.91, whereas the higher evaluation
accuracy gives \textit{Always-Relabel} empirical loss 16. Thus, the calibration estimate leads to
abstention with higher evaluation loss at this cost.

The broader cost grid keeps the same likelihoods and evaluation records while varying the relative
penalties. With $c_{\ActI}=1$, error loss takes values 20, 40, 60, 80, and 100; the
\ActUnresolved{}-to-error ratio is 0.25 or 0.50; and the \ActRelabel{}-to-error ratio is 0.05,
0.075, 0.10, 0.125, 0.15, or 0.20. Across the resulting 60 settings, \textit{JABER} has
lower/equal/higher loss in 18/33/9 settings relative to \textit{Marginal-Projection} and 13/43/4
relative to \textit{CI Updating}. Only six settings reduce loss against both ablations without
increasing erroneous outputs. Changing these relative costs therefore produces ties and losses as
well as gains for \textit{JABER}.

\smallskip\noindent\textbf{Both-passage \ActRelabel{} as a terminal action.}
To examine how gold-based reference selection affects \ActRelabel{}, we let \ActRelabel{} read both
conflicting passages without identifying either as correct. The replay uses the same model,
temperature, and 20/40 gold-NO split, with one additional answer per sample. The passage opposing
gold is always first and the supporting passage second. Each answer is shared by the sample's two
episodes; the initial labels and \ActInspect{} streams are reused. The protocol prespecifies
$c_{\ActR}=1,4$ and reports every integer cost from 1 to 10 under the same \textit{Risk-sensitive}
weights.

\begin{table*}[t]
\centering
\caption{Reference labels, action mappings, and evaluation splits for the public-dataset panels.}
\label{tab:external-diagnostic-overview}
\scriptsize
\setlength{\tabcolsep}{4pt}
\begin{tabularx}{\textwidth}{@{}>{\raggedright\arraybackslash}p{0.12\textwidth}>{\raggedright\arraybackslash}p{0.18\textwidth}>{\raggedright\arraybackslash}X>{\raggedright\arraybackslash}p{0.25\textwidth}@{}}
\toprule
Dataset & Reference label & Action mapping & Data and evaluation split\\
\midrule

\textit{LLMBar} & Human response-pair preference. & Initial choice: \textit{GRM-Llama-3.2-3B}; six further scoring models provide \ActInspect{} judgments. & Model/order selection: 209 training and 82 validation records; test: 98 adversarial and 30 natural pairs.\\
\textit{PPE-IFEval} & Programmatic instruction checks. & Initial model: \textit{InternLM2-7B}; \ActInspect{}: its 20B and 1.8B variants; Blind \ActRelabel{}: cross-family \textit{Gemma}. & 512 instructions, split by instruction: 1,540 development and 520 validation episodes; 495 complete records from 500 held-out episodes.\\
\textit{LLMJudge} & Human relevance ratings. & Initial: \texttt{Olz-gpt4o}; follow-ups: \texttt{Olz-exp}, \texttt{willia-umbrela1}, \texttt{h2oloo-fewself}; replacement: \texttt{prophet-setting4} (\textit{Llama-3-8B}). & 4,423 passages for 25 search questions; five-fold cross-validation grouped by search question.\\
\textit{STEM} expert review & Agreement of the first two experts. & A third independent expert provides an analogue of terminal \ActArbitrate{}. & 150 records; 92 have a third rating, and 76 have agreement between the first two experts.\\
\bottomrule
\end{tabularx}
\end{table*}

Replacement accuracy is 75\% ($15/20$) on calibration and 80\% ($32/40$) on evaluation, giving
$\alpha_{\ActR}=16/22$ (72.73\%) after smoothing. Its calibrated terminal loss,
$c_{\ActR}+80(1-16/22)$, exceeds the \ActUnresolved{} loss of 20 throughout the grid. In the replay,
\textit{JABER} immediately selects \ActUnresolved{} in all 80 episodes. \textit{Always-Relabel} has
empirical loss $c_{\ActR}+16$, which is lower than 20 at $c_{\ActR}=1$--$3$. At these costs, the
lower calibration estimate leads \textit{JABER} to abstain with higher evaluation loss than
\textit{Always-Relabel}. These supplementary comparisons are point estimates from the reused
evaluation records.

\subsection{Additional Diagnostics on Public Datasets}
\label{app:exp-external-details}

This subsection gives the protocols and detailed results for the public-dataset diagnostics,
including those discussed in Section~\ref{subsec:exp-supplementary}, covering repeated-judgment
information, replacement reliability, and expert adjudication.
Table~\ref{tab:external-diagnostic-overview} lists the reference labels, action mappings, and
evaluation splits.

\smallskip\noindent\textbf{Repeated Inspect.}

For \textit{LLMBar}, we compare within- and cross-family joint errors and examine whether follow-up
judgments distinguish correct from incorrect initial choices. Excess joint error is the observed
joint-error probability minus the product of the marginal error probabilities. On the adversarial
test set, this quantity is 5.31 percentage points higher within than across families, with a 95\%
interval of $[2.63,7.96]$. Mean follow-up support is 68.5\% for incorrect and 68.8\% for correct
initial choices; the incorrect-initial-choice subset contains 18 records. Decision replay uses
erroneous-output loss 1, \ActUnresolved{} loss 0.5, and \ActInspect{} costs 0.02 and 0.05. At cost
0.05, additional inspection has higher total loss than \textit{Always-Accept}. These results show
limited discrimination between correct and incorrect initial choices despite the observed error
dependence; at this cost, further inspection is less effective than immediate acceptance.

For \textit{PPE-IFEval}~\cite{frick2025ppe,zhou2023ifeval}, every pair from the same instruction
remains in one split. The development and validation episodes are combined after selecting the
judgment models to fit the final parameters, then evaluated on the complete held-out records listed
in Table~\ref{tab:external-diagnostic-overview}. Same-family models are 18.27 percentage points more
likely than cross-family models to persist with an error, with a 95\% interval of $[7.65,28.61]$.
The probability of persisting twice exceeds the conditional-independence prediction by 4.15
percentage points, with a 95\% interval of $[0.98,7.34]$. The first \ActInspect{} has a $+11.0$
percentage-point support-rate difference between correct and incorrect held choices, whereas the
second has a $-10.5$ percentage-point difference; the fitted model assigns zero evidence weight to
the second step. On the held-out episodes, all three updating methods immediately \ActAccept{} at
all six cost points. These results show persistent errors alongside unstable correctness signals,
limiting the value of repeated inspection under the fitted models and tested costs.

\smallskip\noindent\textbf{Replacement reliability.}

On \textit{LLMJudge}, we compare history-based switching with fixed rules to test whether selective
replacement improves accuracy after three follow-up judgments have been acquired. A label is correct
only if it exactly matches the human 0--3 relevance rating. The switching rule is learned on
training search questions and applied to held-out questions; neither gold labels nor replacement
outcomes from held-out questions are used in learning. Fixed references include always retaining,
always replacing, and taking the mode of the initial and three same-source follow-up labels. This
mode rule adopts the most frequent label, retaining the initial label in a tie.

Across the three follow-up judges, mean support is 17.19 percentage points higher for correct than
incorrect initial labels, with a 95\% interval of $[10.45,24.34]$. The candidate replacement labels,
however, have a higher overall error rate than the initial labels. At zero \ActRelabel{} cost, the
same-source mode rule is the best fixed reference. History-based switching reduces error by 0.723
percentage points relative to this rule, with a 95\% interval of $[-1.094,2.437]$. The interval uses
search-question-level bootstrap resampling, retaining all passages associated with each resampled
question. Thus, selective replacement based on these judgment histories has not shown a clear
accuracy benefit over the fixed mode rule in this configuration.

\smallskip\noindent\textbf{Expert adjudication.}
On the expert-agreement subset in Table~\ref{tab:external-diagnostic-overview}, agreement between
the first two experts supplies the reference label. The third independent expert agrees on 75 of 76
records~\cite{khender2026judgehuman}. Relative to the six LLM judges, the third expert's paired
accuracy is 10.8--25.0 percentage points higher, and all six bootstrap intervals exclude zero. These
comparisons support the reliability of the independent expert on this agreement subset.

\end{document}